\RequirePackage{fix-cm}

\documentclass[smallextended]{svjour3_customheadbox}

\smartqed

\usepackage{graphicx}
\usepackage{geometry}
\usepackage{tabularx}
\usepackage{color}
\usepackage{url}
\usepackage{synttree}
\usepackage{amssymb}
\usepackage{amsmath}
\usepackage{stmaryrd}
\usepackage{tikz}
\usetikzlibrary{calc}

\newcommand{\acetext}[1]{\textsf{#1}}
\newcommand{\qacetext}[1]{``\textsf{#1}''}
\newcommand{\badtext}[1]{\textsf{{\color{gray}*} #1}}

\newcommand{\qcatname}[1]{``{\itshape #1}''}

\newcounter{ruleid}

\newcommand{\possymb}{\ensuremath{\mathord{\#}}}
\newcommand{\scopeopensymb}{\ensuremath{\mathord{\sslash}}}
\newcommand{\fwrefsymb}{\ensuremath{\mathord{>}}}
\newcommand{\sfwrefsymb}{\ensuremath{\mathord{\gg}}}
\newcommand{\bwrefsymb}{\ensuremath{\mathord{<}}}
\newcommand{\nbwrefsymb}{\ensuremath{\mathord{\nless}}}
\newcommand{\cbwrefplus}{\ensuremath{\mathord{^+}}}
\newcommand{\cbwrefminus}{\ensuremath{\mathord{^-}}}
\newcommand{\cbwrefplain}[2]{\ensuremath{\bwrefsymb\cbwrefplus #1 \cbwrefminus #2 }}
\newcommand{\nrulesymb}[0]{\mathrel{:}}
\newcommand{\scrulesymb}[0]{\mathrel{\sim}}
\newcommand{\genrulesymb}[1]{\mathrel{#1}}

\newcommand{\fs}[1]{\!\! \left( \! \scalebox{0.75}{$\begin{array}{l} \\[-2ex] #1 \\[-2ex] \end{array}$} \! \right)}
\newcommand{\nrule}[2]{#1 \: \xrightarrow{\displaystyle \: \nrulesymb \:} \: #2}
\newcommand{\scrule}[2]{#1 \: \xrightarrow{\displaystyle \: \scrulesymb \:} \: #2}
\newcommand{\lrule}[2]{#1 \: \rightarrow \: #2}
\newcommand{\scat}[1]{\:\: \mbox{\itshape #1} \:\:}
\newcommand{\cat}[2]{\:\: \mbox{\itshape #1} \, \fs{#2} }
\newcommand{\fwref}[1]{\:\: \fwrefsymb \fs{#1} }
\newcommand{\sfwref}[1]{\:\: \sfwrefsymb \fs{#1} }
\newcommand{\bwref}[1]{\:\: \bwrefsymb \fs{#1} }
\newcommand{\nbwref}[1]{\:\: \nbwrefsymb \fs{#1} }
\newcommand{\cbwref}[2]{\:\: \bwrefsymb\cbwrefplus #1 \!\! \cbwrefminus #2 }

\newcommand{\term}[1]{\:\: [\,\mbox{#1}\,] \:\:}
\newcommand{\spreterm}[1]{\:\: \mbox{\underline{\itshape #1}} \:\:}
\newcommand{\preterm}[2]{\:\: \mbox{\underline{\itshape #1}} \, \fs{#2} }
\newcommand{\pos}[1]{\:\: \possymb {\fboxsep 0.5mm \framebox{\scalebox{0.6}{#1}}} \:\:}
\newcommand{\scopeopener}[0]{\:\: \scopeopensymb \:\:}

\newcommand{\featv}[2]{\mbox{#1:}\:\fboxsep 0.5mm \framebox{\scalebox{0.8}{#2}}\:\\}
\newcommand{\featc}[2]{\mbox{#1:}\:\mbox{#2}\\}

\newcommand{\edge}[3]{#1 \: \xrightarrow{\displaystyle #2} \: #3}
\newcommand{\edot}[0]{\:\mathord{\bullet}\:}
\newcommand{\epos}[2]{\langle#1,#2\rangle\:\:\:\:}

\newcommand{\tikzpoint}[1]{{\tikz[remember picture] \node (#1) {};}}

\newcommand{\chartpicture}[1]{
  \begin{center}
  \newcommand{\charttransformarrow}{\node at (0,0) (arrow) {\scalebox{3}{$\Rightarrow$}};}
  \scalebox{0.9}{\begin{tikzpicture}[
    mynode/.style={circle,inner sep=.6mm,thick,draw},
    myedge/.style={->,shorten <=0.5mm,shorten >=0.5mm,thick}
  ]
    #1
  \end{tikzpicture}}
  \end{center}
}

\newcommand{\onenodecharttransform}[1]{
  \chartpicture{
    \node[mynode] at (-3,0) (l1)  {};
    \charttransformarrow
    \node[mynode] at (3,0) (r1)  {};
    #1
  }
}

\newcommand{\twonodescharttransform}[1]{
  \chartpicture{
    \node[mynode] at (-4.5,0) (l1)  {};
    \node at (-3.5,0) (tl12)  {$\dots$};
    \node[mynode] at (-2.5,0) (l2)  {};
    \charttransformarrow
    \node[mynode] at (2.5,0) (r1)  {};
    \node at (3.5,0) (tr12)  {$\dots$};
    \node[mynode] at (4.5,0) (r2)  {};
    #1
  }
}

\newcommand{\threenodescharttransform}[1]{
  \chartpicture{
    \node[mynode] at (-5,0) (l1)  {};
    \node at (-4.25,0) (tl12)  {$\dots$};
    \node[mynode] at (-3.5,0) (l2)  {};
    \node at (-2.75,0) (tl23)  {$\dots$};
    \node[mynode] at (-2,0) (l3)  {};
    \charttransformarrow
    \node[mynode] at (2,0) (r1)  {};
    \node at (2.75,0) (tr12)  {$\dots$};
    \node[mynode] at (3.5,0) (r2)  {};
    \node at (4.25,0) (tr23)  {$\dots$};
    \node[mynode] at (5,0) (r3)  {};
    #1
  }
}

\newcommand{\textbox}[1]{\begin{minipage}{6.2cm}\begin{center}\footnotesize #1\end{center}\end{minipage}}
\newcommand{\ndots}{\ensuremath{\ldotp\!\ldotp\!\ldotp}}

\journalname{Journal of Logic, Language and Information}

\begin{document}

\title{
  A Principled Approach to Grammars for Controlled Natural Languages and Predictive Editors
  \thanks{The work presented here was funded by the research grant (Forschungs\-kredit) programs 2006 and 2008 of the University of Zurich. I would like to thank Norbert E. Fuchs, George Herson, Stefan H\"ofler, Michael Hess, Kaarel Kaljurand, Marc Lutz, and Adam Wyner for comments, discussions and proof-reading on this paper and earlier versions thereof.}
}


\author{Tobias Kuhn}


\institute{
  Tobias Kuhn \at
  Department of Pathology, Yale University School of Medicine\\
  \email{kuhntobias@gmail.com}\\
  \texttt{http://www.tkuhn.ch}
}

\date{}

\maketitle

\begin{abstract}
Controlled natural languages (CNL) with a direct mapping to formal logic have been proposed to improve the usability of knowledge representation systems, query interfaces, and formal specifications. Predictive editors are a popular approach to solve the problem that CNLs are easy to read but hard to write. Such predictive editors need to be able to ``look ahead'' in order to show all possible continuations of a given unfinished sentence. Such lookahead features, however, are difficult to implement in a satisfying way with existing grammar frameworks, especially if the CNL supports complex nonlocal structures such as anaphoric references. Here, methods and algorithms are presented for a new grammar notation called Codeco, which is specifically designed for controlled natural languages and predictive editors. A parsing approach for Codeco based on an extended chart parsing algorithm is presented. A large subset of Attempto Controlled English (ACE) has been represented in Codeco. Evaluation of this grammar and the parser implementation shows that the approach is practical, adequate and efficient.
\keywords{Anaphoric references \and Attempto Controlled English \and Chart parsing \and Controlled natural languages \and Predictive editors}
\end{abstract}

\section{Introduction}
\label{sec:introduction}

Controlled natural languages (CNL) \cite{pool2006claw,wyner2009cnlmain} are based on natural language but come with restrictions concerning lexicon, syntax and/or semantics. They have been proposed to make knowledge representations \cite{funk2007iswc,schwitter2008owleddc,fuchs2008reasoningweb,shiffman2009cnlmain}, rule systems \cite{kuhn2007rr,spreeuwenberg2009cnlmain}, query interfaces \cite{mueckstein1985csc,bernstein2006iswc,wuersch2010icse,kuhnhoefler2013coral}, and formal specifications \cite{fuchs1999lopstr} more user-friendly. The scope of this work is restricted to CNLs with a direct and deterministic mapping to some kind of formal logic, like Attempto Controlled English (ACE) \cite{fuchs2008reasoningweb}, Computer Processable English \cite{pulman1999iwcs}, and CLOnE \cite{funk2007iswc}. The approach presented here does \emph{not} target less restricted languages like Basic English \cite{ogden1932basic}, Caterpillar Fundamental English \cite{verbeke1973traindev}, or ALCOGRAM \cite{adriaens1992coling}.

While CNLs are easier to understand than comparable logic languages \cite{kuhn2012swj}, they have the problem that it can be very difficult for users to write statements that comply with the syntactic restrictions of the language.
Three approaches have been proposed so far to solve this problem: error messages \cite{clark2007kcap,dimitrova2008iswc}, conceptual authoring \cite{power2009cnl,franconi2011dl}, and predictive editors \cite{tennant1983acl,schwitter2003eamtclaw,bernstein2006iswc,kuhn2009semwiki}. Each of them has its own advantages and drawbacks. Error messages are the most straightforward approach, and they leave much freedom to users. However, it is very difficult to give good and helpful error messages, because the whole richness of unrestricted language can be expected in statements written by users who are not familiar with the details of the restricted grammar. The other two approaches are based on sophisticated editing environments, which solve the problems of error messages but which also heavily confine the freedom of users during the writing process. Conceptual authoring is a top-down approach that enables users to gradually create concrete statements via given operations on incomplete sentences (i.e. sentences with ``holes''). Predictive editors, in contrast, implement a left-to-right approach. They show all possible continuations of unfinished sentences (i.e. sentences that are ``chopped off'' at the end), enabling step-by-step creation of CNL statements. For any unfinished sentence, a predictive editor shows all possible words or phrases that can be used to continue the sentence in a grammar-compliant way. This approach seems very promising, and it is the one that is followed in the work presented here. Experiments have shown that well-designed predictive editors enable easy creation of well-formed sentences even for users unfamiliar with the respective CNL \cite{bernstein2006iswc,kuhn2009semwiki,kuhnhoefler2013coral}.

Obviously, predictive editors depend on the availability of lookahead features, i.e. the retrieval of the possible continuations to a given unfinished sentence on the basis of a given grammar. This is not a trivial task, especially if the grammar describes complex nonlocal structures like anaphoric references. Apart from that, it is desirable to have a simple grammar notation that is fully declarative and can easily be implemented in different kinds of programming languages. This is because good tool integration is crucial for the usability and acceptance of CNL approaches.

These requirements are explained in more detail in the next section, and existing grammar frameworks are assessed with respect to these requirements (Section \ref{sec:background}). Then, the Codeco grammar notation is presented (Section \ref{sec:codeco}), which is designed to solve the identified problems. Next, an implementation of this grammar notation is introduced in the form of a chart parser (Section \ref{sec:codecocp}), and specific applications of Codeco are discussed (Section \ref{sec:application}), namely a concrete grammar written in Codeco and a CNL editor making use of that grammar. Finally, different aspects of Codeco and its parser are evaluated (Section \ref{sec:evaluation}), before we draw the conclusions (Section \ref{sec:conclusions}).

This paper is a revised and substantially extended version of a workshop paper \cite{kuhn2010cnlmain}. For more details, see the author's doctoral thesis \cite{kuhn2010doctoralthesis}.

\section{Background}
\label{sec:background}

To our knowledge, the Grammatical Framework (GF) \cite{angelov2009cnlmain} is the only existing grammar framework that has a specific focus on CNL. GF fulfills most but not all of the requirements that will be introduced shortly (it has no particular support for describing the resolvability of anaphoric references and does not support dynamic lexica). With extensions (perhaps inspired by Codeco), it could become a suitable grammar notation for the specific problem described in this paper.

The remainder of this section first puts forward a list of grammar requirements (Section \ref{sec:grammarrequ}), and then discusses different kinds of existing grammar frameworks and assesses them with respect to our requirements: natural language grammar frameworks (Section \ref{sec:nlgrammars}), parser generators (Section \ref{sec:parsergen}), and definite clause grammars (Section \ref{sec:dcg}).

\subsection{Grammar Requirements}
\label{sec:grammarrequ}

The implementation of CNLs to be used in predictive editors raises a number of requirements concerning the used grammar notation. We argue for the following six requirements to enable efficient and practical CNL applications:

\paragraph{Concreteness.}
Concreteness is an obvious requirement. Due to their practical and computer-oriented nature, CNL grammars should be concrete, i.e. fully formalized to be read and interpreted by computer programs. With concrete grammars, the syntax trees for given statements can be automatically built, and therefore structural ambiguity can be automatically spotted, among other things.

\paragraph{Declarativeness.}
CNL grammars should be declarative, i.e. defined in a way that does not depend on a specific algorithm or implementation. Declarative grammars can be completely separated from the parser that processes them. This makes it easy to use such grammars from other programs, to replace the parser, or to have different parsers for the same language. Declarative grammars are easy to change and reuse and can be shared easily between different parties using the same CNL.

\paragraph{Lookahead Features.}
Predictive editors require the availability of lookahead features, i.e. the possibility to find out how an unfinished sentence can be continued. For this reason, the CNL should be defined in a form that enables the efficient implementation of such lookahead features. Concretely, this means that an unfinished sentence, for instance \qacetext{a brother of Sue likes ...}, can be given to the parser and that the parser is able return the complete set of words that can be used to continue the sentence according to the grammar. For the given example, the parser might say that \qacetext{a}, \qacetext{every}, \qacetext{no}, \qacetext{somebody}, \qacetext{John}, \qacetext{Sue}, \qacetext{himself} and \qacetext{her} are the possible continuations.

\paragraph{Anaphoric References and Scoping.}
The proper handling of anaphoric references is a tricky issue for predictive editors. Grammars for CNLs that support anaphoric references should describe the circumstances under which such references are allowed in an exact, declarative, and simple way. In order to have a clear separation of syntax and semantics, resolvability of anaphoric references should be treated as a syntactic issue that does not depend on the semantic representation. Concretely, a predictive editor should allow the use of a referential expression like \qacetext{it} only if a matching antecedent (e.g. \qacetext{a country}) can be identified in the preceding text. What makes things more complicated is the fact that the resolvability of anaphoric references depends on the scoping of the preceding text, triggered by negation markers and different kinds of quantifiers. While scoping in natural language can be considered a semantic phenomenon, it has to be treated as a syntactic issue in CNLs if the restrictions on anaphoric references are to be described appropriately. We use the term \emph{anaphoric reference} in a broad sense that includes definite noun phrases like \qacetext{the country}.

\paragraph{Dynamic Lexicon.}
The lexicon of a CNL should be dynamic, i.e. extensible during the parsing process. This is required to make it possible for users to add new words while they are writing a sentence. Being forced to cancel the creation of a sentence just because of a missing word is very frustrating and inefficient.

\paragraph{Implementability.}
Finally, a CNL grammar notation should be simple enough to be easy to implement in different programming languages and should be neutral with respect to the programming paradigm of its parser. This requirement is motivated by the fact that the usability of CNL heavily depends on good integration into user interfaces like predictive editors. For this reason, it is desirable that the CNL parser is implemented in the same programming language as the user interface component. Another reason why implementability is important is the fact that there can be many other tools besides the parser that need to read and process the grammar, including editors, paraphrasers, and verbalizers. Furthermore, more than one parser might be necessary for practical reasons: A simple top-down parser, for example, may be the best option when used for parsing large texts in batch mode and for doing regression tests (e.g. through language generation), but a chart parser is much better suited for providing lookahead capabilities.

\subsection{Natural Language Grammar Frameworks}
\label{sec:nlgrammars}

A large number of grammar frameworks exist for natural languages. Some of the most popular ones are \emph{Head-driven Phrase Structure Grammars} (HPSG) \cite{pollard1994hpsg}, \emph{Lexical-Functional Grammars} \cite{kaplan1982mental}, \emph{Tree-Adjoining Grammars} \cite{joshi1975computsystsci}, \emph{Generalized Phrase Structure Grammars} \cite{gazdar1985gpsg}, and \emph{Combinatory Categorial Grammars} \cite{steedman2011ccg}, but many more exist \cite{cole1997hlt}. Most of these frameworks are defined in an abstract and declarative way. Concrete grammars based on such frameworks, however, are mostly hard-coded in a certain programming language and do not have a declarative nature that would make them independent from their parsers.

Despite many similarities, a number of important differences between natural language grammars and grammars for CNLs can be identified that have the consequence that the grammar frameworks for natural language do not work out very well for CNLs. Most of the differences originate from the fact that the two kinds of grammars are the results of somewhat opposing goals. Natural language grammars are \emph{language descriptions} that describe existing phenomena. CNL grammars, in contrast, are \emph{language definitions} that define new artificial languages, which in some sense just happen to look like natural language.

Obviously, grammars for natural languages and those for CNLs differ in complexity. Natural languages are very complex and so must be the grammars that thoroughly describe such languages. CNLs are typically much simpler and abandon natural structures that are difficult to process.

Partly because of the high degree of complexity, providing lookahead features on the basis of those frameworks is difficult. Another reason is that lookahead seems to be much less relevant for natural language applications, and thus no special attention has been paid to this problem. The difficulty of implementing lookahead features with natural language grammar frameworks can be seen by the fact that no predictive editors exist for CNLs that have emerged from an NLP background like CPL \cite{clark2005flairs} and CLOnE \cite{funk2007iswc} (which are otherwise very different from each other).


The handling of ambiguity is another important difference. Natural language grammars have to deal with the inherent ambiguity of natural language. Context information and background knowledge can help resolving ambiguities, but there is always a remaining degree of uncertainty. Natural language grammar frameworks are designed to be able to cope with such situations, can represent structural ambiguity by using underspecified representations, and require the parser to disambiguate by applying heuristic methods. In contrast, CNLs (the logic-based ones on which this paper focuses) remove ambiguity by their design, which typically makes underspecification and heuristics unnecessary.

Finally, anaphora resolution is another particularly difficult problem for the correct representation of natural language. In computational linguistics, this problem is usually solved by applying algorithms of different degrees of sophistication to find the most likely antecedents (e.g. \cite{hobbs1978lingua,lappin1994coli}). This is a difficult problem because an anaphoric pronoun like \qacetext{it} can refer to a noun phrase that has been introduced in the preceding text but it can also refer to a broader structure like a complete sentence or paragraph. It is also possible that \qacetext{it} refers to something that has been introduced only in an implicit way or to something that will be identified only in the text that follows later. Furthermore, \qacetext{it} can refer to something outside of the text, meaning that background knowledge is needed to resolve the reference. Altogether, this has the consequence that sentences like \qacetext{an object contains it} have to be considered syntactically correct even if no matching antecedent for \qacetext{it} can be clearly identified in the text. This stands in contrast to the requirement introduced in Section \ref{sec:grammarrequ} that a CNL should define the resolvability of anaphoric references on the syntactic level. Natural language grammar frameworks like HPSG establish \emph{binding theories} \cite{chomsky1980inquiry,pollard1994hpsg} to address the problem of anaphoric references. These binding theories consist of principles that describe under which circumstances two components of the text can refer to the same thing. Applying these binding theories, however, just gives a set of possible antecedents for each anaphor and does not deal with deterministic resolution of them.

\subsection{Parser Generators}
\label{sec:parsergen}

A number of systems exist that are aimed at the definition and parsing of formal languages (e.g. programming languages). In the simplest case, such grammars are written in Backus-Naur Form \cite{naur1963acm,knuth1964acm}. Examples of more sophisticated grammar formalisms for formal languages --- called \emph{parser generators} --- include Yacc~\cite{johnson1975yacc} and GNU bison\footnote{\url{http://www.gnu.org/software/bison/}}. Formalized-English \cite{martin2002iccs} is an example of a CNL defined in such a parser generator notation.

The general problem of these formalisms is that context-sensitive constraints cannot be defined in a declarative way. With plain Backus-Naur-style grammars, context-free languages can be described in a declarative and simple way, but such grammars are very limited and even very simple CNLs cannot be defined appropriately. It is possible to include context-sensitive elements, but this has to be done in the form of procedural extensions that depend on a particular programming language to be interpreted. Thus, the property of declarativeness gets lost when it comes to more complex languages.

When discussing lookahead capabilities, it has to be noted that the term \emph{lookahead} has a different meaning in the context of parser generators: \emph{lookahead} denotes how far the parsing algorithm looks ahead in the fixed token list before deciding which rule to apply. Lookahead in our sense of the word --- i.e. predicting possible next tokens --- is not directly supported by existing parser generators. However, as long as no procedural extensions are used, this is not difficult to implement. Actually, a simple kind of lookahead (in our sense of the word) is available in many source code editors in the form of code completion features.

\subsection{Definite Clause Grammars}
\label{sec:dcg}

Definite clause grammars (DCG) \cite{pereira1986nlp} are a simple but powerful notation to define grammars for natural and formal languages. They are almost always written in logic-based programming languages like Prolog. In fact, many of the grammar frameworks for natural languages introduced above are often implemented on the basis of Prolog DCGs.

DCGs are fully declarative in their core (e.g. Prolog DCG rules that do not use curly brackets or cuts). Therefore, they can in principle be processed by any programming language. Building upon the logical concept of definite clauses, they are easy to interpret for logic-based programming languages, but a considerable overhead is necessary in other programming languages to simulate backtracking and unification.

DCGs are good in terms of expressiveness because they are not necessarily context-free but can contain context-sensitive elements. Anaphoric references, however, are again a problem. It is difficult to define them in an appropriate way. The following two exemplary grammar rules show how antecedents and anaphors could be defined:
\begin{quote}\small
\begin{verbatim}
np(Agr, Ante-[Agr|Ante]) --> determiner(Agr), noun(Agr).
np(Agr, Ante-Ante) --> ana_pron(Agr), { once(member(Agr,Ante)) }.
\end{verbatim}
\end{quote}
The code inside the curly brackets unifies the agreement structure of the pronoun with the first possible element of the antecedent list. This part of the code is not fully declarative. A more serious problem, however, is the way connections between anaphors and antecedents are established. The accessible antecedents are passed through the grammar by using input and output lists of the form ``\texttt{In-Out}'' so that new elements can be added to the list whenever an antecedent occurs in the text. The problem that follows from this approach is that the definition of anaphoric references cannot be done locally in the grammar rules that actually deal with anaphoric structures but they affect almost the complete grammar, as illustrated by the following grammar rule:
\begin{quote}\small
\begin{verbatim}
s(Ante1-Ante3) --> np(Agr, Ante1-Ante2), vp(Agr, Ante2-Ante3).
\end{verbatim}
\end{quote}
As this example shows, anaphoric references also have to be considered when writing grammar rules that have otherwise nothing to do with anaphors or antecedents. This is neither convenient nor elegant. This kind of threading of anaphoric information is similar to an approach called \emph{DRS threading} for the generation of semantic representations, in which case anaphoric information can be included with little overhead \cite{johnson1986coling}. However, as motivated above, we would like to treat the resolution of references as a syntactic issue, independent of the semantic representation.

Different DCG extensions have been proposed in order to describe natural language in a more appropriate way. Assumption Grammars~\cite{dahl1997iclp}, for example, are motivated by natural language phenomena that are difficult to express otherwise, like free word order. Simple anaphoric references can be represented in a very clean way, but the approach does not scale up to complex anaphor types like non-reflexive pronouns.

A further problem with the DCG approach concerns lookahead features. In principle, it is possible to provide lookahead features with standard Prolog DCGs \cite{kuhn2008alta}, but this solution is not very efficient and can become impractical for complex grammars and long sentences.

\section{The Codeco Notation}
\label{sec:codeco}

This section presents a new grammar notation called \emph{Codeco}, which stands for ``\emph{co}ncrete and \emph{de}clarative grammar notation for \emph{co}n\-trolled natural languages''. It is specifically designed for CNLs in predictive editors and should allow for relatively simple implementations of complete and correct lookahead features. While many elements of Codeco are straightforward and very similar to existing approaches, the proper handling of anaphora requires novel types of special elements in the form of forward and backward pointing references. With Codeco, anaphoric information flows through the syntax tree in the particular top-down, left-to-right path described by Johnson and Klein \cite{johnson1986coling}.

Below, the elements of the Codeco notation are introduced, i.e. grammar rules, grammatical categories, and certain special elements. These different elements are motivated by ACE examples. After that, the issue of reference resolution is discussed in more detail.

\subsection{Simple Categories and Grammar Rules}

Grammar rules in Codeco use the operator ``$\xrightarrow{\nrulesymb}$'' (where the colon on the arrow is needed to distinguish normal rules from scope-closing rules as they will be introduced later on):
\[\nrule{
  \scat{vp}
}{
  \scat{v}
  \scat{np}
}\]
Terminal categories are represented in square brackets:
\[\nrule{
  \scat{v}
}{
  \term{does not}
  \scat{verb}
}\]
There is a special notation for pre-terminal categories, which are marked with an underline:
\[\nrule{
  \scat{np}
}{
  \term{a}
  \spreterm{noun}
}\]
Pre-terminal categories can be expanded but only to terminal categories, i.e. they can occur on the left hand side of a rule only if the right hand side consists of exactly one terminal category. Such rules are called \emph{lexical rules} and are represented with a plain arrow, for instance:
\[\lrule{
  \spreterm{noun}
}{
  \term{person}
}\]
Lexical rules can be part of the static grammar but they can also be stored in a dynamic lexicon.

In order to support context-sensitivity, non-terminal and pre-terminal categories can be augmented with flat feature structures. Feature representations use the colon operator ``$:$'' with the name of the feature to the left and its value to the right. Values can be variables, which are displayed as boxes:
\[\nrule{
  \cat{vp}{\featv{num}{Num}\featv{neg}{Neg}}
}{
  \cat{v}{\featv{num}{Num}\featv{neg}{Neg}\featc{type}{tr}}
  \cat{np}{\featc{case}{acc}}
}\]
\[\nrule{
  \cat{v}{\featc{neg}{+}\featv{type}{Type}}
}{
  \term{does not}
  \cat{verb}{\featv{type}{Type}}
}\]
\[\nrule{
  \cat{np}{\featv{noun}{Noun}}
}{
  \term{a}
  \preterm{noun}{\featv{text}{Noun}}
}\]
Feature values \emph{cannot} be feature structures themselves, i.e. they are always flat. This restriction has practical reasons. It should keep Codeco simple and easy to implement, but it can easily be dropped in theory.

\subsection{Normal Forward and Backward References}

So far, the introduced elements of Codeco are straightforward and not very specific to CNL or predictive editors. The support for anaphoric references, however, requires some novel extensions. In principle, it is easy to support sentences like
\begin{quote}
  \acetext{Every area that is a part of a stable country is controlled by the country.}
\end{quote}
where \qacetext{the country} is a resolvable anaphoric reference (as mentioned above, our usage of the term \emph{anaphoric reference} includes definite noun phrases). With the Codeco elements introduced so far, however, it is not possible to suppress sentences like
\begin{quote}
  \acetext{Every area is controlled by the country.}
\end{quote}
where \qacetext{the country} is not resolvable. This can be acceptable, but in many situations there are good reasons to disallow such non-resolvable references.

In Codeco, the use of anaphoric references can be restricted to positions where they can be resolved. This is done with the help of the special categories ``\fwrefsymb'' and ``\bwrefsymb'', which describe nonlocal dependencies across the syntax tree, as the following picture shows:
\begin{center}\scalebox{0.8}{
\branchheight{8mm}
\childsidesep{0.3em}
\childattachsep{0.3mm}
\synttree{9}
[s
  [np
    [det
      [.b \acetext{Every}]
    ]
    [n
      [.b \acetext{area}]
    ]
    [$>$]
    [relcl
      [relpron
        [.b \acetext{that}]
      ]
      [vp
        [v
          [cop
            [.b \acetext{is}]
          ]
        ]
        [np
          [det
            [.b \acetext{a}]
          ]
          [n
            [.b \acetext{part}]
          ]
          [$>$]
          [pp
            [prep
              [.b \acetext{of}]
            ]
            [np
              [det
              [.b \acetext{a}]
              ]
              [adj
                [.b \acetext{stable}]
              ]
              [n
                [.b \acetext{country}]
              ]
              [\tikzpoint{n1}]
            ]
          ]
        ]
      ]
    ]
  ]
  [vp
    [aux
      [.b \acetext{is}]
    ]
    [v
      [tv
        [.b \acetext{controlled}]
      ]
    ]
    [pp
      [prep
        [.b \acetext{by}]
      ]
      [np
        [ref
          [det
            [.b \acetext{the}]
          ]
          [n
            [.b \acetext{country}]
          ]
          [\tikzpoint{n2}]
        ]
      ]
    ]
  ]
]
\begin{tikzpicture}[remember picture,overlay]
  \node[circle,inner sep=0.2mm,fill=red!40] at (n1) (p1) {$>$};
  \node[circle,inner sep=0.2mm,fill=red!40] at (n2) (p2) {$<$};
  \draw[very thick,red!30] (p1) .. controls ($(p1) + (2,0)$) and ($(p2) + (-2,0)$) .. (p2);
\end{tikzpicture}
}\end{center}

``\fwrefsymb'' represents a \emph{forward reference} and marks a position in the text to which anaphoric references can refer, i.e. ``\fwrefsymb'' stands for antecedents. ``\bwrefsymb'' represents a \emph{backward reference} and refers back to the closest possible antecedent, i.e. ``\bwrefsymb'' stands for anaphors. These special categories can have feature structures, and they can occur only in the body of rules, for example:
\[\nrule{
  \scat{np}
}{
  \term{a}\preterm{noun}{\featv{text}{Noun}}
  \fwref{\featc{type}{noun}\featv{noun}{Noun}}
}\]
\[\nrule{
  \scat{ref}
}{
  \term{the}\preterm{noun}{\featv{text}{Noun}}
  \bwref{\featc{type}{noun}\featv{noun}{Noun}}
}\]
The forward reference of the first rule establishes an antecedent to which later backward references can refer. The second rule contains such a backward reference that refers back to an antecedent with a matching feature structure. In this example, forward and backward references have to agree in their type and their noun (represented by the features ``type'' and ``noun''). This has the effect that \qacetext{the country}, for example, can refer to \qacetext{a country}, but \qacetext{the area} cannot.

Forward references always succeed, whereas backward references succeed only if a matching antecedent in the form of a forward reference can be found somewhere to the left in the syntax tree. In order to distinguish these simple types of forward and backward references from other reference types that will be introduced below, they are called \emph{normal forward references} and \emph{normal backward references}, respectively.

These special categories provide a very simple way to establish nonlocal dependencies. However, as we will discover, they are not general enough for all types of anaphoric references we would like to represent. We need more reference types, but first accessibility constraints have to be discussed.

\subsection{Scoping and Accessibility}
\label{sec:scopingaccessibility}

As already pointed out, anaphoric references are affected by scoping. References are resolvable only to positions in the previous text that are accessible, i.e. that are not inside closed scopings. An example is
\begin{quote}
  \acetext{Every man protects a house from every enemy and does not destroy ...}
\end{quote}
where one can refer to \qacetext{man} or to \qacetext{house} but not to \qacetext{enemy} (because \qacetext{every} opens a scoping that is closed after \qacetext{enemy}). Therefore, \qacetext{himself} and \qacetext{the house} are possible continuations, but not \qacetext{the enemy}. The Codeco elements introduced so far do not allow for such restrictions. Additional elements are needed to define where scopings open and where they close.

The position where a scoping opens is represented in Codeco by the special category ``\scopeopensymb'' called \emph{scope opener}, for example:
\[\nrule{
  \cat{quant}{\featc{exist}{--}}
}{
  \scopeopener
  \term{every}
}\]

Scopings have to be closed somewhere. In contrast to the opening positions of scopings, their closing positions can be far away from the scope-triggering structure. For this reason, the closing positions cannot be defined in the same way. Instead, Codeco defines scope-closing rules ``$\xrightarrow{\scrulesymb}$'', for instance:
\[\scrule{
  \cat{vp}{\featv{num}{Num}}
}{
  \cat{v}{\featc{neg}{+}\featv{num}{Num}\featc{type}{tr}}
  \cat{np}{\featc{case}{acc}}
}\]
This rule states that any scoping that is opened by the direct or indirect children of \qcatname{v} and \qcatname{np} is closed at the end of \qcatname{np}. If no scopings have been opened, scope-closing rules simply behave like normal rules. See Section \ref{sec:refresprinciples} for a visualization of the scoping and accessibility aspects of the example shown above. In contrast to most other approaches, Codeco defines scopings in a way that is completely independent from the semantic representation.

\subsection{Position Operators}

With the introduced Codeco elements, anaphoric definite noun phrases like \qacetext{the area} can be restricted to positions where they are resolvable. At this point, however, we cannot define that a reflexive pronoun like \qacetext{herself} is allowed only if it refers to the subject of the respective verb phrase. Concretely, we cannot distinguish the following two cases:
\begin{quote}
  \acetext{A woman helps herself.}\\
  \badtext{A woman knows a man who helps herself.}
\end{quote}
The problem is that there is no way to check whether a potential antecedent is the subject of a given anaphoric reference or not. What is needed is a way of assigning an identifier to each antecedent.

To this aim, Codeco employs the position operator ``\possymb'', which takes a variable and assigns it an identifier that represents the respective position in the text. The following picture visualizes how position operators work (each $p_i$ being a position identifier):
\begin{center}\scalebox{0.8}{
\branchheight{8mm}
\childsidesep{1em}
\childattachsep{0.3mm}
\synttree{5}
[s
  [np
    [\tikzpoint{n1}]
    [prop
      [.b \tikzpoint{l1}\acetext{Mary}\tikzpoint{r1}]
    ]
  ]
  [vp
    [v
      [tv
        [.b \tikzpoint{l2}\acetext{helps}\tikzpoint{r2}]
      ]
    ]
    [np
      [\tikzpoint{n2}]
      [ref
        [.b \tikzpoint{l3}\acetext{herself}\tikzpoint{r3}]
      ]
    ]
  ]
]
\begin{tikzpicture}[remember picture,overlay]
  \node[circle,inner sep=0.5mm,fill=blue!40,draw=black] at ($(l1) + (-0.6,0)$) (p0) {\scriptsize $p_0$};
  \node[circle,inner sep=0.5mm,draw=black] at ($0.5*($(r1) + (l2)$)$) (p1) {\scriptsize $p_1$};
  \node[circle,inner sep=0.5mm,fill=blue!40,draw=black] at ($0.5*($(r2) + (l3)$)$) (p2) {\scriptsize $p_2$};
  \node[circle,inner sep=0.5mm,draw=black] at ($(r3) + (0.6,0)$) (p3) {\scriptsize $p_3$};
  \draw[very thick,blue!40] (n1) -| (p0);
  \draw[very thick,blue!40] (n2) -| (p2);
  \node[circle,inner sep=0.2mm,fill=blue!40] at (n1) {$\#$};
  \node[circle,inner sep=0.2mm,fill=blue!40] at (n2) {$\#$};
\end{tikzpicture}
}\end{center}

With the use of position operators, reflexive pronouns can be defined in a way that excludes unresolvable pronouns, i.e. excludes pronouns that do not match with the subject of the given verb phrase:
\[\nrule{
  \cat{np}{\featv{id}{Id}}
}{
  \pos{Id}
  \cat{prop}{\featv{human}{H}}
  \fwref{\featv{id}{Id}\featv{human}{H}\featc{type}{prop}}
}\]
\[\nrule{
  \cat{ref}{\featv{subj}{Subj}}
}{
  \term{itself}
  \bwref{\featv{id}{Subj}\featc{human}{--}}
}\]
As we will see, a further extension is needed for the appropriate definition of non-reflexive pronouns.

\subsection{Negative Backward References}

We need to solve a further problem, which concerns variables as they are supported by some CNLs like ACE. Phrases like \qacetext{a person X} can be used to introduce a variable \qacetext{X}. The question is how to treat cases where the same variable is introduced twice:
\begin{quote}
  \badtext{A person X knows a person X.}
\end{quote}
One solution is to allow such sentences and to define that the second introduction of \qacetext{X} overrides the first one, such that subsequent occurrences of \qacetext{X} can only refer to the second but not to the first. In first-order logic, for example, variables are treated this way. In CNL, however, the overriding of variables can be confusing to the readers. ACE, for example, does not allow variables to be overridden.

Such restrictions cannot be defined with the Codeco elements introduced so far. Another extension is needed: the special category ``\nbwrefsymb'', which ensures that there is no matching antecedent. This special category establishes \emph{negative backward references}, which can be used --- among other things --- to ensure that no variable is introduced twice:
\[\nrule{
  \scat{newvar}
}{
  \preterm{var}{\featv{text}{V}}
  \nbwref{\featc{type}{var}\featv{var}{V}}
  \fwref{\featc{type}{var}\featv{var}{V}}
}\]
The special category ``\nbwrefsymb'' succeeds only if there is no accessible forward reference that unifies with the given feature structure.

\subsection{Complex Backward References}

The Codeco elements that have been introduced are still not sufficient for expressing all the things we would like to express. As already mentioned, there is still a problem with non-reflexive pronouns like \qacetext{him}. While reflexive pronouns like \qacetext{himself} can be restricted to refer only to the respective subject, non-reflexive pronouns cannot be prevented from doing so:
\begin{quote}
  \acetext{John knows Bill and helps him.}\\
  \badtext{John helps him.}
\end{quote}
To distinguish such cases, it becomes necessary to introduce \emph{complex backward references}, which use the special structure ``\cbwrefplain{\ndots}{\ndots}''. Complex backward references can have several feature structures: one or more positive ones (after the symbol ``\cbwrefplus''), which define how a matching antecedent must look like, and zero or more negative ones (after ``\cbwrefminus''), which define how the antecedent must \emph{not} look like. The symbol ``\cbwrefminus'' can be omitted if no negative feature structures are present. This allows for correctly representing non-reflexive pronouns:
\[\nrule{
  \cat{ref}{\featv{subj}{Subj}}
}{
  \term{he}
  \cbwref{ \fs{\featc{human}{+}\featc{gender}{masc}} }{ \fs{\featv{id}{Subj}} }
}\]
Complex backward references refer to the closest accessible forward reference that unifies with one of the positive feature structures but is not unifiable with any of the negative ones.

Complex backward references are powerful constructs, which restrict anaphoric references in a very general way. The following example --- which is rather artificial and would probably not be very useful in practice --- illustrates the general nature of complex backward references: Say that \qacetext{this} should be used to refer to antecedents which are either neuter and have no variable attached or which are of type ``relation'' (whatever that means), while in neither case being a proper name or the subject of the sentence. This complex behavior could be achieved with the following rule:
\[\nrule{
  \cat{ref}{\featv{subj}{Subj}}
}{
  \term{this}
  \cbwref{ \fs{\featc{hasvar}{--}\featc{human}{--}} \fs{\featc{type}{relation}} }{ \fs{\featc{type}{prop}} \fs{\featv{id}{Subj}} }
}\]
Complex backward references that have exactly one positive feature structure and no negative ones are equivalent to normal backward references.

\subsection{Strong Forward References}

Finally, one last extension is needed in order to handle antecedents that are not affected by the accessibility constraints. Usually, proper names are considered accessible even if under negation:
\begin{quote}
  \acetext{Mary does not love Bill. Mary hates him.}
\end{quote}
In such situations, the special category ``\sfwrefsymb'' can be used, which introduces a \emph{strong forward reference}:
\[\nrule{
  \cat{np}{\featv{id}{Id}}
}{
  \cat{prop}{\featv{human}{H}}
  \sfwref{\featv{id}{Id}\featv{human}{H}\featc{type}{prop}}
}\]
Strong forward references are always accessible even if they are within closed scopings. Apart from that, they behave like normal forward references.

\subsection{Principles of Reference Resolution}
\label{sec:refresprinciples}

The resolution of references in Codeco requires some more explanation. All three types of backward references (normal, negative and complex ones) are resolved according to the three principles of accessibility, proximity and left-dependence.

\paragraph{Accessibility.}

The principle of accessibility states that one can refer to forward references only if they are accessible from the position of the backward reference. A forward reference is accessible only if it is not within a scoping that has been closed before the position of the backward reference, or if it is a strong forward reference.

This accessibility constraint can be visualized in the syntax tree. For the unfinished sentence shown in Section~\ref{sec:scopingaccessibility}, the syntax tree could look as follows:
\begin{center}\scalebox{0.75}{
\branchheight{8mm}
\childsidesep{0.3em}
\childattachsep{0.3mm}
\synttree{8}
[s $\sim$
  [np
    [det
      [\tikzpoint{s1}]
      [.b \tikzpoint{l1}\acetext{Every}]
    ]
    [n
      [.b \acetext{man}]
    ]
    [\tikzpoint{fw1}]
  ]
  [vp
    [vp \tikzpoint{t1}
      [v
        [tv
          [.b \acetext{protects}]
        ]
      ]
      [np
        [det
          [.b \acetext{a}]
        ]
        [n
          [.b \acetext{house}]
        ]
        [\hspace{1mm} \tikzpoint{fw2}\hspace{1mm} ]
      ]
      [pp
        [prep
          [.b \acetext{from}]
        ]
        [np
          [det
            [\tikzpoint{s2}]
            [.b \tikzpoint{l2}\acetext{every}]
          ]
          [n
            [.b \acetext{enemy}]
          ]
          [$>$]
        ]
      ]
    ]
    [conj
      [.b \tikzpoint{r1}\acetext{and}]
    ]
    [vp $\sim$
      [aux
        [\tikzpoint{s3}]
        [.b \tikzpoint{l3}\acetext{does not}]
      ]
      [v
        [tv
          [.b \acetext{destroy}]
        ]
      ]
      [np
        [ref
          [.b \acetext{...}]
          [\hspace{1mm} \tikzpoint{bw1}\hspace{1mm} ]
        ]
      ]
    ]
  ]
]
\begin{tikzpicture}[remember picture,overlay]
  \node at ($(l1) + (-0.2,0)$) (p1) {};
  \node at ($(l2) + (-0.2,0)$) (p2) {};
  \node at ($(l3) + (-0.2,0)$) (p3) {};
  \node at ($(r1) + (-0.15,0)$) (p4) {};
  \fill[fill opacity=0.1,blue] (t1 -| p2) rectangle (p4.south);
  \draw[very thick,blue!40] (s1) -| (p1);
  \draw[very thick,blue!40] (s2) -| (p2);
  \draw[very thick,blue!40] (s3) -| (p3);
  \draw[very thick,blue!40] (t1) -| (p4);
  \node[circle,inner sep=0.2mm,fill=blue!40] at (t1) (t1n) {$\sim$};
  \node[circle,inner sep=0.2mm,fill=blue!40] at (s1) (s1n) {$\scopeopensymb$};
  \node[circle,inner sep=0.2mm,fill=blue!40] at (s2) (s2n) {$\scopeopensymb$};
  \node[circle,inner sep=0.2mm,fill=blue!40] at (s3) (s3n) {$\scopeopensymb$};
  \node[circle,inner sep=0.5mm,fill=blue!40,draw=black] at (p1) (p1n) {\scriptsize $($};
  \node[circle,inner sep=0.5mm,fill=blue!40,draw=black] at (p2) (p2n) {\scriptsize $($};
  \node[circle,inner sep=0.5mm,fill=blue!40,draw=black] at (p3) (p3n) {\scriptsize $($};
  \node[circle,inner sep=0.5mm,fill=blue!40,draw=black] at (p4) (p4n) {\scriptsize $)$};
  
  \node[circle,inner sep=0.2mm,fill=red!40] at (fw1) (fw1n) {$>$};
  \node[circle,inner sep=0.2mm,fill=red!40] at (fw2) (fw2n) {$>$};
  \node[circle,inner sep=0.2mm,fill=red!40] at (bw1) (bw1n) {$<$};
  \draw[very thick,red!40,dashed] (fw1n) .. controls ($(fw1n) + (2,0)$) and ($(bw1n) + (-2,0)$) .. (bw1n);
  \draw[very thick,red!40,dashed] (fw2n) .. controls ($(fw2n) + (3,0)$) and ($(bw1n) + (-3,0)$) .. (bw1n);
\end{tikzpicture}
}\end{center}
All nodes that represent the head of a scope-closing grammar rule are marked with ``$\sim$''. The positions in the text where scopings open and close are marked with parentheses. In this example, three scopings have been opened but only the second one (the one in front of \qacetext{every enemy}) has been closed (after \qacetext{enemy}). The shaded area marks the part of the syntax tree that is covered by this closed scoping. As a consequence of the accessibility constraint, the forward references for \qacetext{man} and \qacetext{house} are accessible from the position of the backward reference at the very end of the shown unfinished sentence. In contrast, the forward reference for \qacetext{enemy} is not accessible because it is inside a closed scoping. The possible references are shown as dashed lines. Thus, the unfinished sentence can be continued with the anaphoric references \qacetext{the man} or \qacetext{the house} (or equivalently \qacetext{himself} or \qacetext{it}, respectively) but not with the reference \qacetext{the enemy}.

\paragraph{Proximity.}

Proximity is the second principle for the resolution of backward references. If a backward reference could potentially point to more than one forward reference then, as a last resort, the principle of proximity defines that the textually closest forward reference is taken. More precisely, when traversing the syntax tree starting from the backward reference and going back in a right-to-left, depth-first manner, the first matching and accessible forward reference is taken. This ensures that every backward reference resolves deterministically to exactly one forward reference.

In the following example, the reference \qacetext{it} could in principle refer to three antecedents:
\begin{center}\scalebox{0.75}{
\branchheight{8mm}
\childsidesep{0.3em}
\childattachsep{0.3mm}
\synttree{7}
[s
  [conj
  	[.b \acetext{If}]
  ]
  [s
    [np
      [det
        [.b \acetext{a}]
      ]
      [n
        [.b \acetext{part}]
      ]
      [\hspace{1mm} \tikzpoint{fw1}\hspace{1mm} ]
      [pp
        [prep
          [.b \acetext{of}]
        ]
        [np
          [det
            [.b \acetext{a}]
          ]
          [n
            [.b \acetext{machine}]
          ]
          [\hspace{1mm} \tikzpoint{fw2}\hspace{1mm} ]
        ]
      ]
    ]
    [vp
      [v
        [tv
          [.b \acetext{causes}]
        ]
      ]
      [np
        [det
          [.b \acetext{an}]
        ]
        [n
          [.b \acetext{error}]
        ]
        [\hspace{1mm} \tikzpoint{fw3}\hspace{1mm} ]
      ]
    ]
  ]
  [conj
  	[.b \acetext{then}]
  ]
  [s
    [np
      [ref
        [pn
          [.b \acetext{it}]
        ]
        [\hspace{1mm} \tikzpoint{bw1}\hspace{1mm} ]
      ]
    ]
    [\acetext{...}
      [.b \acetext{...}]
    ]
  ]
]
\begin{tikzpicture}[remember picture,overlay]
  \node[circle,inner sep=0.2mm,fill=red!40] at (fw1) (fw1n) {\fwrefsymb};
  \node[circle,inner sep=0.2mm,fill=red!40] at (fw2) (fw2n) {\fwrefsymb};
  \node[circle,inner sep=0.2mm,fill=red!40] at (fw3) (fw3n) {\fwrefsymb};
  \node[circle,inner sep=0.2mm,fill=red!40] at (bw1) (bw1n) {\bwrefsymb};
  \draw[very thick,red!40,dashed] (fw1n) .. controls ($(fw1n) + (6,0)$) and ($(bw1n) + (-0.5,0)$) .. (bw1n);
  \draw[very thick,red!40,dashed] (fw2n) .. controls ($(fw2n) + (4,0)$) and ($(bw1n) + (-0.5,0)$) .. (bw1n);
  \draw[very thick,red!40] (fw3n) .. controls ($(fw3n) + (1,0)$) and ($(bw1n) + (-1,0)$) .. (bw1n);
\end{tikzpicture}
}\end{center}
The pronoun \qacetext{it} could refer to \qacetext{part}, \qacetext{machine}, or \qacetext{error}. According to the principle of proximity, the closest antecedent is taken, i.e. \qacetext{error}.

\paragraph{Left-dependence.}

The principle of left-dependence, finally, means that everything to the left of a backward reference is considered for its resolution but everything to its right is not. The crucial point is that variable bindings entailed by a part of the syntax tree to the left of the reference are considered, whereas variable bindings that would be entailed by a part of the syntax tree to the right are not considered.

The following example illustrates why the principle of left-dependence is important:
\[\nrule{
  \scat{ref}
}{
  \term{the}
  \bwref{\featc{type}{noun}\featv{noun}{N}}
  \preterm{noun}{\featv{text}{N}}
}\]
\[\nrule{
  \scat{ref}
}{
  \term{the}
  \preterm{noun}{\featv{text}{N}}
  \bwref{\featc{type}{noun}\featv{noun}{N}}
}\]
These are two versions of the same grammar rule. The only difference is that the backward reference and the pre-terminal category \qcatname{noun} are switched. The first version is not a very sensible one: the backward reference is resolved without considering how the variable ``N'' is bound by the category \qcatname{noun}. The second version is much better: the resolution of the reference takes into account which noun has been read from the input text.

As a rule of thumb, backward references should generally follow the textual representation of the anaphoric reference and not precede it.

\subsection{Restriction on Backward References}

In order to provide proper and efficient lookahead algorithms that can handle backward references, their usage must be restricted: Backward references must immediately follow a terminal or pre-terminal category in the body of grammar rules. Thus, they are not allowed at the initial position of the rule body and they must not follow a non-terminal category. This restriction is important for the lookahead algorithm to be presented, but it is not relevant for the parsing algorithm.

\section{Codeco in a Chart Parser}
\label{sec:codecocp}

In order to provide lookahead features for predictive editors, chart parsers are a good choice. In addition, they are well-suited for implementations in procedural or object-oriented programming languages, because they do not depend on backtracking. The basic idea is to store temporary parse results in a data structure that is called a \emph{chart} and that contains small portions of parse results in the form of \emph{edges}.

The algorithm for Codeco to be presented here is based on the chart parsing algorithm invented by Jay Earley, which is therefore known as the \emph{Earley algorithm} \cite{earley1970acm}. Grune and Jacobs \cite{grune2008parsing} discuss this algorithm in more detail, and Covington \cite{covington1994nlpprolog} shows how it can be implemented. The specialty of the Earley algorithm is that it combines top-down and bottom-up processing.

The parsing time of the standard Earley algorithm is in the worst case cubic with respect to the number of tokens to be parsed and only quadratic for the case of unambiguous grammars. However, this holds only if the categories have no arguments (e.g. feature structures). Otherwise, parsing is NP-complete in the general case. This means that for \emph{certain} grammars, longer sentences cannot be parsed efficiently. Fortunately, grammars describing natural language typically do not fall into this worst-case category. Furthermore, there is a certain soft upper limit on the length of natural sentences (sentences of more than 1000 words are rarely found in a natural context, and only a few instances with more than 10\,000 words are reported). For these reasons, Earley parsers usually perform very well for natural grammars and the majority of input texts.

The algorithm described in this section has been implemented in Java. This implementation is the basis for the predictive editor that is used in AceWiki and the ACE Editor, which will be introduced in the next section. The code is available as open source as part of the AceWiki code base.\footnote{\url{https://github.com/AceWiki/AceWiki}}

To describe the elements of the chart parser and to explain the different steps of the parsing algorithm, the following meta language symbols will be used:
\begin{quote}
\begin{description}
\item[$F$] stands for a feature structure, i.e. a set of name/value pairs.
\item[$A$] stands for any (terminal, pre-terminal or non-terminal) category, i.e. a category name followed by an optional feature structure.
\item[$\alpha$] stands for an arbitrary sequence of zero or more categories.
\item[$r$] stands for a forward reference symbol, i.e. either ``\fwrefsymb'' or ``\sfwrefsymb''.
\item[$\rho$] stands for an arbitrary sequence of zero or more forward references ``$rF$'' and scope openers ``\scopeopensymb''.
\item[$s$] stands for either a colon ``$\nrulesymb$'' or a tilde ``$\scrulesymb$'' so that ``$\xrightarrow{s}$'' can stand for ``$\xrightarrow{\nrulesymb}$'' or for ``$\xrightarrow{\scrulesymb}$''.
\item[$i$] stands for a position identifier that represents a certain position in the input text.
\end{description}
\end{quote}
All meta symbols can have a numerical index to distinguish different instances of the same symbol, e.g. $\alpha_1$ and $\alpha_2$. Other meta symbols will be introduced as needed.

Below, the chart parser elements and parsing steps are explained (Sections \ref{sec:cpelements} and \ref{sec:cpsteps}), before the actual lookahead algorithm is introduced (Section \ref{sec:codecolookahead}).

\subsection{Chart Parser Elements}
\label{sec:cpelements}

Before we can turn to the actual parsing steps, the fundamental elements of Earley parsers are introduced (chart and edges) together with a graphical notation that will be used to describe the parsing steps in an intuitive way.

\paragraph{Edges.}

Every edge of a chart parser is derived from a grammar rule and --- like the grammar rule --- consists of a head and a body:
\[
  \epos{i_1}{i_2} \edge{A}{}{\alpha_1 \edot \alpha_2}
\]
The body of the edge is split by the dot symbol ``$\bullet$'' into a sequence $\alpha_1$ of categories that have already been recognized and another sequence $\alpha_2$ of categories that still have to be processed. In addition, every edge has a start position $i_1$ and an end position $i_2$. Such an edge tells us that we started looking for category $A$ at position $i_1$. Up to position $i_2$, we have found the category sequence $\alpha_1$, but to complete category $A$, we still need to recognize the sequence $\alpha_2$.

Edges where all categories of the body are recognized are called \emph{passive}. All other edges are called \emph{active}, and their first category of the sequence of not-yet-recognized categories is called their \emph{active category}.

\paragraph{Edges with Antecedents.}

Processing Codeco grammars requires an extended notation for edges. Whenever a backward reference occurs, we need to be able to find out which antecedents are accessible from that position. For this reason, edges coming from a Codeco grammar have to carry information about the accessible antecedents.

First of all, every edge must carry the information whether it originated from a normal rule or a scope-closing one. Edges originating from normal rules are called \emph{normal edges} and edges coming from scope-closing rules are called \emph{scope-closing edges}. Like the rules they originate from, normal edges are represented by an arrow with a colon ``$\xrightarrow{\nrulesymb}$'' and scope-closing edges use an arrow with a tilde ``$\xrightarrow{\scrulesymb}$''.

In addition, every Codeco edge has two sequences which are called \emph{external antecedent list} and \emph{internal antecedent list}. Both lists are displayed above the arrow: the external one to the left of the colon or tilde, the internal one to the right thereof. Both antecedent lists are sequences of forward references and scope openers. Hence, Codeco edges have the following general structure:
\[
  \epos{i_1}{i_2} \edge{A}{\rho_1 \genrulesymb{s} \rho_2}{\alpha_1 \edot \alpha_2}
\]
$\rho_1$ is the external antecedent list. It represents the antecedents that come from outside the edge, i.e. from positions earlier than $i_1$. $\rho_2$ is the internal antecedent list. It contains the antecedents that come from inside the edge, i.e. from the categories of $\alpha_1$ and their children. Internal antecedents come from somewhere between the start and the end position of the respective edge. Scope openers in the antecedent lists show where scopings have been opened that are not yet closed up to the given position.

\paragraph{Chart.}

A chart is a data structure used to store the partial parse results in the form of edges. During the parsing process, edges are added to the chart, which is initially empty. Edges are not changed or removed from the chart once they are added (unless the input text changes). Traditionally, chart parsers perform a \emph{subsumption check} for each new edge to be added to the chart~\cite{covington1994nlpprolog}: A new edge is added only if no equivalent or more general edge already exists. For reasons that will become clear later, the algorithm to be presented requires an \emph{equivalence check} instead of a subsumption check: New edges are added to the chart except for the case that there exists an edge that is fully equivalent.

\paragraph{Graphical Notation.}

In order to be able to describe the chart parsing steps for the Codeco notation in an intuitive way, a simple graphical notation is used that is inspired by Gazdar and Mellish~\cite{gazdar1989prolognlp}. The positions of the input text are represented by small circles that are arranged as a horizontal sequence, and edges are represented as arrows that point from their start position to their end position having a label with the remaining edge information.
This gives the following representation of a general edge:
\chartpicture{
  \node[mynode] at (0,0) (n1) {};
  \node at (1,0) (tr12)  {$\dots$};
  \node[mynode] at (2,0) (n2) {};
  \draw[myedge] (n1) to [bend left=45] (n2);
  \node[above of=tr12] {$\edge{A}{\rho_1 \genrulesymb{s} \rho_2}{\alpha_1 \edot \alpha_2}$};
  \node[anchor=north] at (n1.south) {$i_1$};
  \node[anchor=north] at (n2.south) {$i_2$};
}
The three dots ``$\dots$'' mean that $i_2$ is either the same position as $i_1$ or directly follows $i_1$ or indirectly follows $i_1$. Thus, it includes the case $i_1 = i_2$ that would be represented as
\chartpicture{
  \node[mynode] at (0,0) (n1) {};
  \draw[myedge] (n1) to [loop above] (n1);
  \node[above of=n1] {$\edge{A}{\rho_1 \genrulesymb{s} \rho_2}{\alpha_1 \edot \alpha_2}$};
  \node[anchor=north] at (n1.south) {$i_1 = i_2$};
}
Active edges can be generally represented as
\chartpicture{
  \node[mynode] at (0,0) (n1) {};
  \node at (1,0) (tr12)  {$\dots$};
  \node[mynode] at (2,0) (n2) {};
  \draw[myedge] (n1) to [bend left=45] (n2);
  \node[above of=tr12] {$\edge{A_1}{\rho_1 \genrulesymb{s} \rho_2}{\alpha_1 \edot A_2\alpha_2}$};
  \node[anchor=north] at (n1.south) {$i_1$};
  \node[anchor=north] at (n2.south) {$i_2$};
}
where $A_2$ is the active category of the edge. Passive edges, in contrast, have the general form
\chartpicture{
  \node[mynode] at (0,0) (n1) {};
  \node at (1,0) (tr12)  {$\dots$};
  \node[mynode] at (2,0) (n2) {};
  \draw[myedge] (n1) to [bend left=45] (n2);
  \node[above of=tr12] {$\edge{A}{\rho_1 \genrulesymb{s} \rho_2}{\alpha \edot}$};
  \node[anchor=north] at (n1.south) {$i_1$};
  \node[anchor=north] at (n2.south) {$i_2$};
}
where the dot is at the last position of the body. This graphical notation is used below to describe the parsing steps in an explicit but intuitive way.

\subsection{Chart Parsing Steps}
\label{sec:cpsteps}

In a traditional Earley parser, there are four parsing steps: initialization, scanning, prediction and completion. In the case of Codeco, an additional step --- to be called \emph{resolution} --- is needed to resolve the references, position operators, and scope openers. Below, the general algorithm is explained, each of the five parsing steps is described, and some brief complexity considerations are given.

\subsubsection{General Algorithm}

The general algorithm starts with the initialization step to initialize the empty chart. Then, prediction, completion and resolution are performed several times, which together will be called the \emph{PCR} step. This PCR step corresponds to the ``Completer/Predictor Loop'' as described by Grune and Jacobs~\cite{grune2008parsing}. A text is parsed by consecutively scanning the tokens of the text and by performing the PCR step after each scanning of a token. The following piece of pseudocode shows this general algorithm:
\begin{quote}\small
\begin{verbatim}
parse(tokens) {
    new chart
    initialize(chart)
    pcr(chart)
    foreach t in tokens {
        scan(chart,t)
        pcr(chart)
    }
}
\end{verbatim}
\end{quote}
The PCR step consists of repeatedly executing the prediction, completion and resolution steps until none of the three is able to generate an edge that is not yet in the chart:
\begin{quote}\small
\begin{verbatim}
pcr(chart) {
    loop {
        c := chart.size()
        predict(chart)
        complete(chart)
        resolve(chart)
        if c=chart.size() then return
    }
}
\end{verbatim}
\end{quote}
The actual order of these three steps can be changed without breaking the algorithm, but it can have an effect on the performance.

In terms of performance, there is potential for optimization anyway. First of all, the above algorithm checks the chart for new edges only after the resolution step. An optimized algorithm checks after each step whether the last three steps contributed a new edge or not. Furthermore, a progress table can be introduced that allows the different parsing steps to remember which edges of the chart they already checked. In this way, edges can be prevented from being checked by the same parsing step more than once. Such an optimized algorithm can look as follows (without going into the details of the progress table):
\begin{quote}\small
\begin{verbatim}
pcr(chart) {
    step := 0
    i := 0
    new progressTable
    loop {
        c := chart.size()
        if step=0 then predict(chart,progressTable)
        if step=1 then complete(chart,progressTable)
        if step=2 then resolve(chart,progressTable)
        if c=chart.size() then i := i+1 else i := 0
        if i>2 then return
        step := (step+1) modulo 3
    }
}
\end{verbatim}
\end{quote}
The variable \texttt{i} counts the number of consecutive idle steps, i.e. steps that did not increase the number of edges in the chart. The loop can be exited as soon as this value reaches $3$. In this situation, no further edge can be added because each of the three sub-steps has been performed on exactly the same chart without being able to add a new edge.

\subsubsection{Graphical Notation for Parsing Steps}

Building upon the graphical notation for edges introduced above, the parsing steps will be described by the use of a graphical notation with a large arrow in the middle of the picture corresponding to the following scheme:
\chartpicture{
  \charttransformarrow
  \node at (-3.5,0) {(edges in the chart)};
  \node at (3.5,0) {(edge to be added)};
  \node at (0,-0.5) {(rule in the grammar)};
}
On the left hand side of the arrow, edges are shown that need to be in the chart in order to execute the described parsing step. If a grammar rule is shown below the arrow then this rule must be present in the grammar for executing the parsing step. On the right hand side of the arrow the new edge is shown that has to be added to the chart when the described parsing step is executed, unless the resulting edge is already there.

If a certain meta symbol occurs more than once on the left hand side of the picture and in the rule representation below the arrow then this means that the respective parts have to be unifiable but not necessarily identical. When generating the new edge, these unifications have to be considered but the existing edges in the chart and the grammar rules remain unchanged.

\subsubsection{Initialization}

At the very beginning, the chart has to be initialized. For each rule that has the topmost category (something like \qcatname{text} or \qcatname{sentence}) on its left hand side, an edge is introduced into the chart at the start position:
\onenodecharttransform{
  \draw[myedge] (r1) to [loop above] (r1);
  \node at (0,-0.5) {$\edge{I}{\genrulesymb{s}}{\alpha}$};
  \node[above of=r1] {$\edge{I}{\genrulesymb{s}}{\edot \alpha}$};
  \node[anchor=north] at (l1.south) {$i_0$};
  \node[anchor=north] at (r1.south) {$i_0$};
}
$i_0$ stands for the start position, i.e. the position in front of the first token, and $I$ stands for the topmost category of the grammar. The only difference to the standard Earley algorithm is that the information about normal and scope-opening rules is taken over from the grammar to the chart, represented by $s$.

\subsubsection{Scanning}

During the scanning step, a token is read from the input text. This token is interpreted as a terminal symbol $T$, for which a passive edge is introduced that has $T$ on its left hand side:
\chartpicture{
  \node[mynode] at (-4.5,0) (l1)  {};
  \node at (-3.5,0) (tl12)  {$T$};
  \node[mynode] at (-2.5,0) (l2)  {};
  \charttransformarrow
  \node[mynode] at (2.5,0) (r1)  {};
  \node at (3.5,0) (tr12)  {$T$};
  \node[mynode] at (4.5,0) (r2)  {};
  \draw[myedge] (r1) to [bend left=45] (r2);
  \node[above of=tr12] {$\edge{T}{\nrulesymb}{\edot}$};
}
In addition, the token can occur in one or more lexical rules:
\chartpicture{
  \node[mynode] at (-4.5,0) (l1)  {};
  \node at (-3.5,0) (tl12)  {$T$};
  \node[mynode] at (-2.5,0) (l2)  {};
  \charttransformarrow
  \node[mynode] at (2.5,0) (r1)  {};
  \node at (3.5,0) (tr12)  {$T$};
  \node[mynode] at (4.5,0) (r2)  {};
  \draw[myedge] (r1) to [bend left=45] (r2);
  \node[above of=tr12] {$\edge{P}{\nrulesymb}{T \edot}$};
  \node at (0,-0.5) {$\lrule{P}{T}$};
}
The rule $\lrule{P}{T}$ can come from the static grammar or from a dynamic lexicon.

\subsubsection{Prediction}

The prediction step looks out for grammar rules that could be applied at the given position. For every active category in the chart that matches the head of a rule in the grammar, a new edge is created:
\twonodescharttransform{
  \draw[myedge] (l1) to [bend left=45] (l2);
  \draw[myedge] (r2) to [loop above] (r2);
  \node[above of=tl12] {$\edge{A}{\rho_1 \genrulesymb{s_1} \rho_2}{\alpha_1 \edot N \alpha_2}$};
  \node at (0,-0.5) {$\edge{N}{\genrulesymb{s_2}}{\alpha_3}$};
  \node[above of=r2] {$\edge{N}{\rho_1 \rho_2 \genrulesymb{s_2}}{\edot \alpha_3}$};
}
$N$ denotes a non-terminal category. The external antecedent list of the new edge is a concatenation of the external and the internal antecedent list of the existing edge. The internal antecedent list of the new edge is empty because it has no recognized categories in its body and thus cannot have internal antecedents.

Because the two shown nodes can actually refer to the same position, the new edge that is produced by such a prediction step can itself be used to produce more new edges by again applying the prediction step at the same position.

\subsubsection{Completion}

The completion step takes the active categories of the active edges in the chart and looks for passive edges with a corresponding head. If such two edges can be found then a new edge can be created out of them.

In the standard Earley algorithm, there is only one kind of completion step. In the case of Codeco, however, it is necessary to differentiate between the cases where the passive edge is a normal edge and those where it is a scope-closing edge. In the case of a normal edge, the completion step looks as follows:
\threenodescharttransform{
  \draw[myedge] (l1) to [bend left=45] (l2);
  \draw[myedge] (l2) to [bend right=45] (l3);
  \draw[myedge] (r1) to [bend left=35] (r3);
  \node[above of=tl12] {$\edge{A_1}{\rho_1 \genrulesymb{s} \rho_2}{\alpha_1 \edot A_2 \alpha_2}$};
  \node[below of=tl23] {$\edge{A_2}{\rho_1 \rho_2 \nrulesymb \rho_3}{\alpha_3 \edot}$};
  \node[above of=r2] {$\edge{A_1}{\rho_1 \genrulesymb{s} \rho_2 \rho_3}{\alpha_1 A_2 \edot \alpha_2}$};
}
In contrast to the standard Earley algorithm, not only the active category of the active edge has to match the head of the passive edge, but also the references of the active edge have to be present in the same order in the passive edge.

If no scoping has been opened then scope-closing edges are completed in exactly the same way as normal edges:
\threenodescharttransform{
  \draw[myedge] (l1) to [bend left=45] (l2);
  \draw[myedge] (l2) to [bend right=45] (l3);
  \draw[myedge] (r1) to [bend left=35] (r3);
  \node[above of=tl12] {$\edge{A_1}{\rho_1 \genrulesymb{s} \rho_2}{\alpha_1 \edot A_2 \alpha_2}$};
  \node[below of=tl23] {$\edge{A_2}{\rho_1 \rho_2 \scrulesymb \rho_3}{\alpha_3 \edot}$};
  \node[above of=r2] {$\edge{A_1}{\rho_1 \genrulesymb{s} \rho_2 \rho_3}{\alpha_1 A_2 \edot \alpha_2}$};
  \node[below of=l2,anchor=north] {\textbox{\vspace{7mm}where $\rho_3$ contains no scope opener \scopeopensymb}};
}
If one or more scopings have been opened then everything that comes after the first scope opener (except strong forward references) is removed from the internal antecedent list for the new edge to be added:
\threenodescharttransform{
  \draw[myedge] (l1) to [bend left=45] (l2);
  \draw[myedge] (l2) to [bend right=45] (l3);
  \draw[myedge] (r1) to [bend left=35] (r3);
  \node[above of=tl12] {$\edge{A_1}{\rho_1 \genrulesymb{s} \rho_2}{\alpha_1 \edot A_2 \alpha_2}$};
  \node[below of=tl23] {$\edge{A_2}{\rho_1 \rho_2 \scrulesymb \rho_3 \scopeopener \rho_4}{\alpha_3 \edot}$};
  \node[above of=r2] {$\edge{A_1}{\rho_1 \genrulesymb{s} \rho_2 \rho_3 \rho_5}{\alpha_1 A_2 \edot \alpha_2}$};
  \node[below of=l2,anchor=north] {\textbox{\vspace{7mm}where $\rho_3$ contains no scope opener \scopeopensymb}};
  \node[below of=r2,anchor=north] {\textbox{where $\rho_5$ is the sequence of all $\sfwrefsymb$-references that appear in $\rho_4$ (in the same order)}};
}
This rule ensures that references within scopings are not accessible from the outside.

\subsubsection{Resolution}

In order to handle position operators, scope openers, and references, an additional parsing step is needed, which is called \emph{resolution}. Generally, only elements occurring in the position of an active category are resolvable.

A position operator is resolved by unifying its variable with an identifier that represents the given position in the input text:
\twonodescharttransform{
  \draw[myedge] (l1) to [bend left=45] (l2);
  \draw[myedge] (r1) to [bend left=45] (r2);
  \node[above of=tl12] {$\edge{A}{\rho_1 \genrulesymb{s} \rho_2}{\alpha_1 \edot \possymb i \; \alpha_2}$};
  \node[above of=tr12] {$\edge{A}{\rho_1 \genrulesymb{s} \rho_2}{\alpha_1 \, \possymb i \edot \alpha_2}$};
  \node[anchor=north] at (l2.south) {$i$};
}

Scope openers are resolved by adding the scope opener symbol to the end of the internal antecedent list:
\twonodescharttransform{
  \draw[myedge] (l1) to [bend left=45] (l2);
  \draw[myedge] (r1) to [bend left=45] (r2);
  \node[above of=tl12] {$\edge{A}{\rho_1 \genrulesymb{s} \rho_2}{\alpha_1 \edot \scopeopensymb \, \alpha_2}$};
  \node[above of=tr12] {$\edge{A}{\rho_1 \genrulesymb{s} \rho_2 \scopeopensymb}{\alpha_1 \, \scopeopensymb \edot \alpha_2}$};
}

Forward references are resolved in a similar way. Together with their feature structure, they are added to the end of the internal antecedent list:
\twonodescharttransform{
  \draw[myedge] (l1) to [bend left=45] (l2);
  \draw[myedge] (r1) to [bend left=45] (r2);
  \node[above of=tl12] {$\edge{A}{\rho_1 \genrulesymb{s} \rho_2}{\alpha_1 \edot \fwrefsymb F \, \alpha_2}$};
  \node[above of=tr12] {$\edge{A}{\rho_1 \genrulesymb{s} \rho_2 \fwrefsymb F}{\alpha_1 \, \fwrefsymb F \edot \alpha_2}$};
}
\twonodescharttransform{
  \draw[myedge] (l1) to [bend left=45] (l2);
  \draw[myedge] (r1) to [bend left=45] (r2);
  \node[above of=tl12] {$\edge{A}{\rho_1 \genrulesymb{s} \rho_2}{\alpha_1 \edot \sfwrefsymb F \, \alpha_2}$};
  \node[above of=tr12] {$\edge{A}{\rho_1 \genrulesymb{s} \rho_2 \sfwrefsymb F}{\alpha_1 \, \sfwrefsymb F \edot \alpha_2}$};
}

Complex backward references can be resolved to an internal antecedent or --- if this is not possible --- to an external one. The resolution to an internal antecedent works as follows (with $1 \leq x \leq m$ and $0 \leq n$):
\twonodescharttransform{
  \draw[myedge] (l1) to [bend left=45] (l2);
  \draw[myedge] (r1) to [bend right=45] (r2);
  \node[above of=tl12] {$\edge{A}{\rho_1 \genrulesymb{s} \rho_2 r F_1 \rho_3}{\alpha_1 \edot \cbwrefplain{F'_1 \ndots F'_x \ndots F'_m}{F''_1 \ndots F''_n} \, \alpha_2}$};
  \node[below of=tr12] {$\edge{A}{\rho_1 \genrulesymb{s} \rho_2 r F_2 \rho_3}{\alpha_1 \, \cbwrefplain{F'_1 \ndots F_2 \ndots F'_m}{F''_1 \ndots F''_n} \edot \alpha_2}$};
  \node[below of=tl12,yshift=-8mm,anchor=north] {\textbox{where $F_1$ is unifiable with $F'_x$ and is not unifiable with any $F''$, and where $\rho_3$ contains no $rF_3$ such that $F_3$ is unifiable with an $F'$ while being not unifiable with any $F''$}};
  \node[below of=tr12,yshift=-8mm,anchor=north] {\textbox{where $F_1$ and $F'_x$ are unified and $F_2$ is the result of this unification}};
}
The positive feature structures of the complex backward reference are denoted by $F'$, the negative ones by $F''$. The resolution to an external antecedent is straightforward:
\twonodescharttransform{
  \draw[myedge] (l1) to [bend left=45] (l2);
  \draw[myedge] (r1) to [bend right=45] (r2);
  \node[above of=tl12] {$\edge{A}{\rho_1 r F_1 \rho_2 \genrulesymb{s} \rho_3}{\alpha_1 \edot \cbwrefplain{F'_1 \ndots F'_x \ndots F'_m}{F''_1 \ndots F''_n} \, \alpha_2}$};
  \node[below of=tr12] {$\edge{A}{\rho_1 r F_2 \rho_2 \genrulesymb{s} \rho_3}{\alpha_1 \, \cbwrefplain{F'_1 \ndots F_2 \ndots F'_m}{F''_1 \ndots F''_n} \edot \alpha_2}$};
  \node[below of=tl12,yshift=-8mm,anchor=north] {\textbox{where $F_1$ is unifiable with $F'_x$ and is not unifiable with any $F''$, and where $\rho_2$ and $\rho_3$ contain no $rF_3$ such that $F_3$ is unifiable with an $F'$ while being not unifiable with any $F''$}};
  \node[below of=tr12,yshift=-8mm,anchor=north] {\textbox{where $F_1$ and $F'_x$ are unified and $F_2$ is the result of this unification}};
}
Note that the same edge can produce more than one new edge when several positive feature structures can unify with the same forward reference. Since normal backward references are equivalent to complex ones for the case $x = m = 1$ and $n = 0$, they do not need to be discussed separately.

Negative backward references, finally, can be resolved only if no matching antecedent exists, neither internal nor external:
\twonodescharttransform{
  \draw[myedge] (l1) to [bend left=45] (l2);
  \draw[myedge] (r1) to [bend left=45] (r2);
  \node[above of=tl12] {$\edge{A}{\rho_1 \genrulesymb{s} \rho_2}{\alpha_1 \edot \nbwrefsymb F_1 \, \alpha_2}$};
  \node[above of=tr12] {$\edge{A}{\rho_1 \genrulesymb{s} \rho_2}{\alpha_1 \, \nbwrefsymb F_1 \edot \alpha_2}$};
  \node[below of=tl12,anchor=north] {\textbox{where $\rho_1$ and $\rho_2$ contain no $rF_2$ such that $F_2$ can unify with $F_1$}};
}
Here it becomes clear why an equivalence check --- and not just a subsumption check --- is needed before adding new edges to the chart. A negative backward reference that is resolvable given certain antecedent lists is not necessarily resolvable in the case of antecedent lists that are more general. More specific edges can behave differently than general ones, and for this reason an edge has to be added to the chart even if a more general edge already exists.

\subsubsection{Complexity Considerations}

Here, some brief and scruffy complexity considerations for the presented algorithm are given. This is done by comparing it to the standard Earley algorithm, which has been proven to be efficient in practical applications.

The space requirements of chart parsers can be measured by the size of the chart, i.e. by the number of contained edges. As long as the positive feature structures of complex backward references are pairwise disjoint (i.e. not unifiable), the special elements of Codeco increase the number of edges in the chart --- compared to the standard Earley algorithm --- only linearly with respect to the number of special elements used in the grammar, and only by a constant factor with respect to the length of the token list. This can be seen by the fact that the scanning, prediction, and completion steps do not produce more edges than in the standard algorithm. Furthermore, for each edge and its descendant edges that contain special elements, the resolution step can be applied at most once for each special element. The only exception are complex backward references with positive feature structures that are not pairwise disjoint. Thus, the chart can be expected to remain reasonably small as long as complex backward references with more than one positive feature structure are used with caution.

In terms of time complexity, it is easy to verify that the additional time needed --- compared to the standard algorithm --- for processing any edge in the prediction or resolution step or any two edges in the completion step is linear with respect to the number of elements in the external and internal antecedent list. Since the number of elements in the antecedent lists is linearly correlated with the number of parsed tokens and since the number of tokens increases the chart only by a constant factor, it can be concluded that the amount of additional time that is needed grows only in a linear way with respect to the number of tokens. Furthermore, checking for the equivalence of edges does not take more than twice as much time compared to checking for subsumption (because equivalence can be checked by a mutual check for subsumption). Altogether, the presented algorithm can be expected to be reasonably fast.

\subsection{Lookahead with Codeco}
\label{sec:codecolookahead}

Given Codeco's chart parsing algorithm, lookahead features --- as they are needed for predictive editors --- can be efficiently implemented in a relatively simple way. The basic idea is that the lookahead information is stored in the active categories. These are categories that are predicted to possibly occur after the end position of the respective edge. Thus, the possible next tokens can be found in the active terminal categories of the edges that have their end positions at the end of the given unfinished sentence. Pre-terminal categories and backward references, however, make the actual algorithm a little more complicated.

The possible next tokens are described as sets of options where at least one of the options must be fulfilled by a token to be a possible continuation of a given unfinished sentence. The algorithm to be introduced can describe the possible next tokens in an abstract and in a concrete way by generating a set of abstract options $O_a$ and another set of concrete options $O_c$. An abstract option would say, for example, that any proper name is a possible next token, whereas a concrete option could say that the concrete proper name \qacetext{Bill} is a possible token.

In order to get this lookahead information, the unfinished sentence has to be parsed, i.e. the chart has to be filled with the edges that represent the syntactic structure of the unfinished sentence. As a next step, the abstract options can be extracted. After that, the concrete options can be created using the abstract options and the lexicon entries.

\paragraph{Extraction of Abstract Options.}

First of all, a formal structure for abstract options has to be defined. In the algorithm to be presented, abstract options have the form
\[
  C / \left\{X_1 \ldots X_n\right\}
\]
with $n \geq 0$ and where $C$ and each $X_j$ are terminal or pre-terminal categories. $C$ denotes a category of possible next tokens with $X_j$ being exceptions in the form of more specific categories describing tokens that are not possible. For instance, the abstract option
\[
  \spreterm{var} / \left\{\preterm{var}{\featc{varname}{X}}\preterm{var}{\featc{varname}{Z}}\right\}
\]
states that all tokens of the pre-terminal category \qcatname{var} are possible next tokens with the exception of those with a ``varname'' feature value of ``X'' or ``Z''. Concretely, this means that any variable is a possible next token except \qacetext{X} and \qacetext{Z}. Another example is
\[
  \preterm{pron}{\featc{refl}{--}\featc{gender}{fem}} / \left\{\right\}
\]
that denotes that any non-reflexive feminine pronoun is a possible next token. Terminal categories can also appear in abstract options, e.g.
\[
  \term{that} / \left\{\right\}
\]
stating that the word \qacetext{that} is a possible next token.

The set of abstract options $O_a$ is extracted from the edges of the chart. This is done by iterating over all edges that have their end position at the position where the unfinished sentence ends. This position is denoted by $i_x$. Only edges are relevant that have a terminal or pre-terminal category (denoted by $T$) as their active category.

First, let us consider edges that have a complex backward reference after their active category. For every edge --- and for every possible $F'_x$ therein --- of the form
\chartpicture{
  \node[mynode] at (0,0) (n1) {};
  \node at (1,0) (tr12)  {$\dots$};
  \node[mynode] at (2,0) (n2) {};
  \draw[myedge] (n1) to [bend left=45] (n2);
  \node[above of=tr12] {$\edge{A}{\rho_1 \genrulesymb{s} \rho_2}{\alpha_1 \edot T \cbwrefplain{F'_1 \ndots F'_x \ndots F'_m}{F''_1 \ndots F''_n} \alpha_2}$};
  \node[anchor=north] at (n2.south) {$i_x$};
}
with $1 \leq x \leq m$ and $0 \leq n$, and for every $rF_1$ that is contained in $\rho_1$ or in $\rho_2$ and that has a feature structure $F_1$ that is unifiable with $F'_x$, an abstract option
\[
  T' / \left\{T''_1 \ldots T''_t\right\}
\]
is added to $O_a$ where $T'$ is the result of category $T$ after unifying $F_1$ and $F'_x$ and where the exceptions are obtained as follows: For every $F''$ that is unifiable with $F_1$, an exception $T''$ is added that is the result of category $T$ after unifying $F_1$ and $F''$. The differentiation between $T$ and $T'$ is necessary because the unification of $F_1$ and $F'_x$ can entail the binding of variables that also occur in $T$. Altogether, this has the effect that terminal or pre-terminal categories in front of backward references are reported as possible next tokens with exceptions that describe all cases for which the reference can afterwards not be resolved. Again, this part of the algorithm described on the basis of complex backward references also applies for normal backward references, which will not be discussed separately because they are just a special case.

Next, we have to handle edges with negative backward references. For every edge of the form
\chartpicture{
  \node[mynode] at (0,0) (n1) {};
  \node at (1,0) (tr12)  {$\dots$};
  \node[mynode] at (2,0) (n2) {};
  \draw[myedge] (n1) to [bend left=45] (n2);
  \node[above of=tr12] {$\edge{A}{\rho_1 \genrulesymb{s} \rho_2}{\alpha_1 \edot T \nbwrefsymb F \alpha_2}$};
  \node[anchor=north] at (n2.south) {$i_x$};
}
an abstract option
\[
  T / \left\{T'_1 \ldots T'_n\right\}
\]
is added to $O_a$ where the exceptions $T'_i$ are obtained as follows: For every $rF'$ that is contained in $\rho_1$ or in $\rho_2$ and that has a feature structure $F'$ that is unifiable with $F$, an exception $T'_j$ is added that is the result of category $T$ after unifying $F$ and $F'$. The effect of this is that terminal or pre-terminal categories that are in front of negative backward references are reported as possible next tokens together with exceptions that describe all cases where the negative backward reference can afterwards find a matching antecedent and thus cannot be resolved.

So far, these option descriptions ``look'' only one step ahead. They do not cover cases where more than one terminal or pre-terminal category exists between the active position and the backward reference. In the case of normal and complex backward references, however, it is possible and useful to look more than one step ahead. The symbol $\delta$ is used to represent a sequence of one or more terminal or pre-terminal categories.

For every edge --- and for every possible $F'_x$ therein --- of the form
\chartpicture{
  \node[mynode] at (0,0) (n1) {};
  \node at (1,0) (tr12)  {$\dots$};
  \node[mynode] at (2,0) (n2) {};
  \draw[myedge] (n1) to [bend left=45] (n2);
  \node[above of=tr12] {$\edge{A}{\rho_1 \genrulesymb{s} \rho_2}{\alpha_1 \edot T \delta \cbwrefplain{F'_1 \ndots F'_x \ndots F'_m}{F''_1 \ndots F''_n} \alpha_2}$};
  \node[anchor=north] at (n2.south) {$i_x$};
}
and for every $rF$ that is contained in $\rho_1$ or in $\rho_2$ and that has a feature structure $F$ that is unifiable with $F'_x$, an abstract option
\[
  T' / \left\{\right\}
\]
is added to $O_a$ where $T'$ is the result of category $T$ after unifying $F$ and $F'_x$.

Finally, edges that have a terminal or pre-terminal category at their active position but are not covered by the patterns introduced so far, an abstract option is created the following way: For every edge of the form
\chartpicture{
  \node[mynode] at (0,0) (n1) {};
  \node at (1,0) (tr12)  {$\dots$};
  \node[mynode] at (2,0) (n2) {};
  \draw[myedge] (n1) to [bend left=45] (n2);
  \node[above of=tr12] {$\edge{A}{\rho_1 \genrulesymb{s} \rho_2}{\alpha_1 \edot T \alpha_2}$};
  \node[anchor=north] at (n2.south) {$i_x$};
}
that is not covered by the patterns introduced above, an abstract option
\[
  T / \left\{\right\}
\]
is added to $O_a$. This means that when no backward reference is close to the active category then this terminal or pre-terminal category is reported as a category of a possible next token.

In this way, a set of abstract options $O_a$ is obtained that describes the possible next tokens in a general way, i.e. without considering the lexicon. Such general lookahead information can be important for predictive editors, e.g. for allowing users to add new words on the fly.

\paragraph{Extraction of Concrete Options.}

In contrast to abstract options, which describe possible next tokens without explicitly listing them, concrete options show the concrete terminal categories that are possible at the given position in the text.

Concrete options could actually just be terminal categories. For user-friendly predictive editors, however, it can be necessary to know the pre-terminal categories from which they are derived, e.g. for grouping the possible next words into different submenus. For this reason, concrete options have the form
\[
  W \leftarrow C
\]
where $W$ is a terminal category and $C$ is a pre-terminal category from which $W$ has been derived. The following example represents the possibility to continue the unfinished text with the noun \qacetext{country}:
\[
  \term{country} \leftarrow \preterm{noun}{\featc{human}{--}}
\]

The special symbol ``$\varnothing$'' is used at the position of $C$ if the given word does not originate from a lexical rule but directly from the grammar. The concrete option
\[
  \term{every} \leftarrow \varnothing
\]
for instance, states that \qacetext{every} is a possible next token that does not come from the lexicon but is part of the grammar rules.

The set of concrete options $O_c$ is generated on the basis of the abstract options $O_a$. For every abstract option
\[
  W / \left\{\right\}
\]
that is contained in $O_a$ and where $W$ is a terminal category, a concrete option
\[
  W \leftarrow \varnothing
\]
is added to $O_c$. In addition, for every abstract option
\[
  C / \left\{X_1 \ldots X_n\right\}
\]
where $C$ is a pre-terminal category, and for each lexical rule
\[\lrule{
  C'
}{
  W
}\]
where $C'$ is unifiable with $C$ but is not unifiable with any $X_j$, the concrete option
\[
  W \leftarrow C
\]
is added to $O_c$.

In this way, we obtain a set of concrete options $O_c$ containing the concrete word forms that are possible to follow the given unfinished sentence.

\paragraph{Lookahead Interface.}

Parsers that implement these algorithms can provide a simple interface for predictive editors to access the lookahead features. Assuming that the unfinished text has been submitted to the parser, the predictive editor module can simply request the set of concrete options $O_c$ and --- if needed --- the set of abstract options $O_a$. In this way, the predictive editor module has all needed information in order to show how the text can be continued, e.g. in the form of graphical menus.

The set of concrete options can directly be presented to the user as possible next words. On the basis of the set of abstract options, the predictive editor can, for example, allow users to create new words that are not yet known at the time the lookahead algorithm runs. Thus, the predictive editor does not only know which concrete words are possible at the given position but also which words in general would be allowed if they were in the lexicon.

This kind of lookahead is not necessarily restricted to the level of individual words. Tokens can be multi-word units such as \qacetext{it is false that}. Moreover, the introduced lookahead algorithm can be applied several times to find sequences of two, three or more tokens to continue the given unfinished sentence. A predictive editor could show a selection of these, e.g. the most frequently used ones. The predictive editor implementation to be introduced shortly does not currently support this kind of multi-token lookahead, but there are no technical obstacles to implementing it with the presented techniques.

\paragraph{Completeness and Correctness.}

The presented lookahead algorithm can be analyzed with respect to completeness and correctness: It is complete in the sense that it returns every token for which, together with the tokens of the unfinished sentence, a \emph{complete} syntax tree exists that is well-formed according to the grammar. It is correct in the sense that it only returns the tokens for which, together with the tokens of the unfinished sentence, a \emph{partial} syntax tree exists that is well-formed and does not end with an unresolvable reference.

These definitions of completeness and correctness leave some freedom on how to handle certain special cases. They do not say anything about the tokens that lead to a well-formed partial syntax tree that cannot be completed to a full statement. This can happen, for example, if the edge used for predicting the next token contains at a later position a non-terminal category that does not occur as a head in any of the grammar rules. In this case, the edge can never complete and the predicted token is actually not a possible next token to complete the unfinished sentence, even though a well-formed partial syntax tree can be constructed.

A stronger correctness criterion that requires the existence of a completion for the partial syntax tree seems not practical, because its detection would entail a complex search problem on the grammar rules. Thus, Codeco grammars should be designed in such a way that invalid statements fail at the earliest possible position, i.e. at the first position for which no continuation to a well-formed statement exists. It could be argued that properly designed grammars should follow this restriction anyway.

\section{Applications}
\label{sec:application}

\begin{figure}[t]
\begin{center}
\newcommand{\component}[6]{
  \fill[top color=#1!40!white!90!black,bottom color=#1!20,fill,rounded corners] (#2,#3) rectangle ($(#2,#3) + (#4,#5)$);
  \draw[color=#1!30!black,rounded corners,thick] (#2,#3) rectangle ($(#2,#3) + (#4,#5)$);
  \node at ($(#2,#3) + 0.5*(#4,0) + 0.5*(0,#5)$) {\begin{minipage}{#4cm}\begin{center}\footnotesize\sffamily #6\end{center}\end{minipage}};
}
\begin{tikzpicture}
\component{white}{0}{2}{12.2}{0.8}{\textbf{ACE Editor}\\\scriptsize(concrete, executable, stand-alone)}
\component{white}{6.2}{1}{6}{0.8}{\textbf{Codeco Chart Parser}\\\scriptsize(concrete, executable, not stand-alone)}
\component{white}{0}{1}{6}{0.8}{\textbf{ACE Codeco Grammar}\\\scriptsize(concrete, not executable)}
\component{white}{0}{0}{12.2}{0.8}{\textbf{Codeco Notation}\\\scriptsize(abstract)}
\end{tikzpicture}
\caption{The dependencies of the different components (higher components depend on lower ones). The attributes in parentheses emphasize their different types.}
\label{fig:aceeditorarch}
\end{center}
\end{figure}
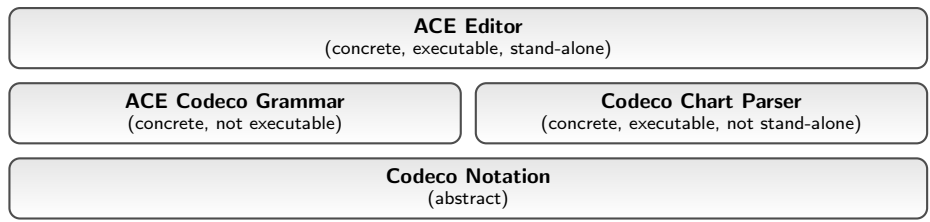

This section briefly introduces two concrete applications of the Codeco notation: the ACE Codeco grammar and the ACE Editor. As illustrated in Figure \ref{fig:aceeditorarch}, these two applications are on different conceptual levels. The ACE Editor depends on the ACE Codeco grammar and on the Codeco chart parser, which all depend on the Codeco notation.

\subsection{ACE Codeco Grammar}
\label{sec:acecodeco}

The ACE Codeco grammar is --- as its name suggests --- a grammar in the Codeco notation describing a subset of the language ACE. It consists of 164 grammar rules\footnote{See \cite{kuhn2010doctoralthesis} for the complete grammar.} and covers a large part of ACE including countable nouns, proper names, intransitive and transitive verbs, adjectives, adverbs, prepositions, plurals, negation, comparative and superlative adjectives and adverbs, \emph{of}-phrases, relative clauses, modality, numerical quantifiers, coordination of sentences / verb phrases / relative clauses, conditional sentences, and questions. Anaphoric references are supported in the form of simple definite noun phrases, variables, and reflexive and non-reflexive pronouns. However, there are some considerable restrictions with respect to the full language of ACE: Mass nouns, measurement nouns, ditransitive verbs, numbers and strings as noun phrases, sentences as verb phrase complements, Saxon genitive, possessive pronouns, noun phrase coordination, and commands are not covered at this point.

Nevertheless, this subset of ACE defined by the Codeco grammar is probably the broadest unambiguous subset of English that has ever been defined in a concrete and fully declarative way and that includes complex nonlocal phenomena like anaphoric references.

\subsection{ACE Editor}

\begin{figure}[t]
\begin{center}
\includegraphics[width=\textwidth]{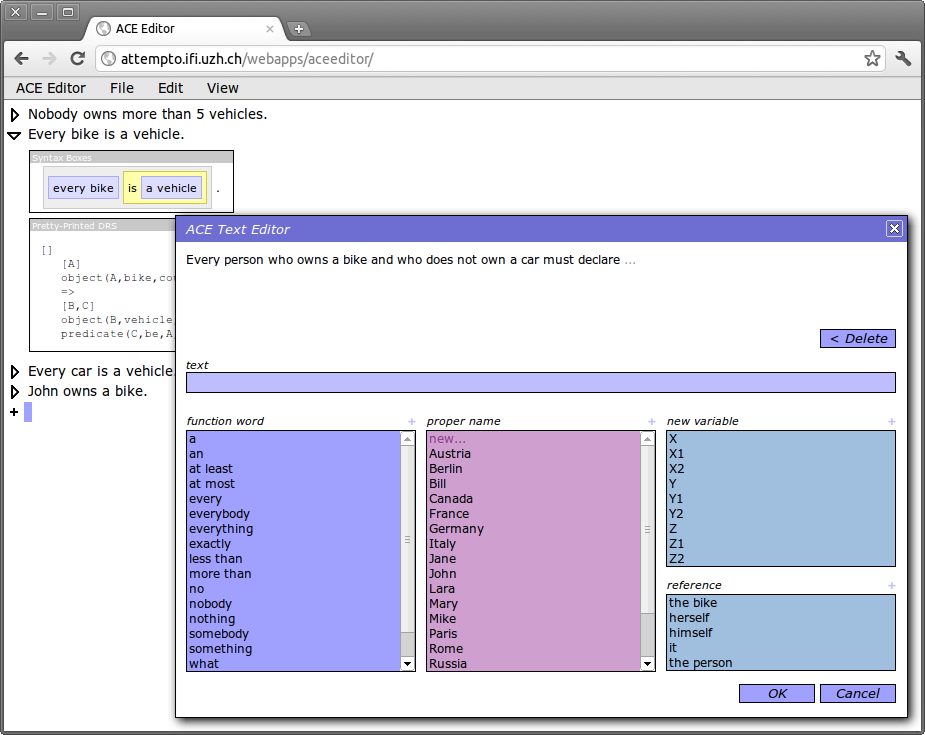}
\caption{The ACE Editor with its predictive editor.}
\label{fig:aceeditor}
\end{center}
\end{figure}

The ACE Editor\footnote{\url{http://attempto.ifi.uzh.ch/webapps/aceeditor/}} is a general web-based editor for writing and modifying ACE texts. Its purpose is to demonstrate how user interfaces can be designed and used to write and modify CNL texts in a simple and intuitive way. Users should not need to learn the grammar of ACE in advance, but they should be able to learn the language while using the editor. A predictive editor is included that uses the ACE Codeco grammar introduced above. Different representations for ACE sentences can be shown like syntax trees, paraphrases and logical formulas, which are all generated by the ACE parser. Figure~\ref{fig:aceeditor} shows a screenshot with the predictive editor as an internal window.
The predictive editor is a modular component that can be reused by other applications. In fact, AceWiki \cite{kuhn2008swui,kuhn2009semwiki} and Coral \cite{kuhnhoefler2013coral} use this predictive editor. No reasoning features are implemented at this point, but applications like AceWiki demonstrate how reasoning can be tightly integrated in such systems.

Obviously, the predictive editor is the most important part of the ACE Editor. In designing predictive editors, the biggest challenge is arguably the fact that there can be a large number of possible ways to continue an unfinished sentence. All these possibilities should be shown to the users in a simple and understandable way, allowing them to quickly choose the option they are looking for. To solve this problem, the ACE Editor uses menu boxes that occupy most of the space of the predictive editor window. Each of these boxes contains the menu items for a particular type of word, showing scroll bars when the given vertical space is not sufficient. The text field above the menu boxes can be used to filter the items. In this way, large numbers of possible next tokens can be presented to the users, and the selection of an item only requires one mouse click and possibly some scrolling or filtering.

The unfinished sentence shown in the predictive editor in Figure~\ref{fig:aceeditor} demonstrates the capability of Codeco to correctly predict resolvable anaphoric references. For the given unfinished sentence, the editor suggests \qacetext{the bike} and \qacetext{the person} but not \qacetext{the car}, because \qacetext{car} is under the scope of the negation expressed with \qacetext{does not} and is therefore not accessible.

\section{Evaluation}
\label{sec:evaluation}

Different aspects of the approach have to be evaluated. Basically, all four components shown in Figure \ref{fig:aceeditorarch} need scrutiny. The top component has been evaluated in our previous work: The usability of the ACE Editor's predictive editor has been tested in several end user experiments. The results showed that it is effectively usable by untrained participants \cite{kuhn2009semwiki,kuhnhoefler2013coral}. Below, evaluation results on the two middle components are presented: the ACE Codeco grammar and the chart parser. The bottom component --- the Codeco notation --- is an abstract one and does therefore not allow for direct evaluation, but we can get indirect results from evaluating the concrete grammar and the parser.

The evaluation presented here relies on the method of exhaustive language generation, i.e. the generation of all possible sentences from a given grammar and lexicon up to a certain fixed sentence length. The main problem of this approach is that one quickly encounters combinatorial explosion on the number of generated sentences. In practice, this means that one can only use a subset of the grammar. Such an evaluation subset has been defined for ACE Codeco, using a minimal lexicon and only 97 of the 164 grammar rules. These 97 grammar rules are chosen in a way that reduces combinatorial explosion but retains the complexity of the language. The lexicon of this subset only contains one entry per word category: the proper name \qacetext{Mary}, the noun \qacetext{woman}, the adjective \qacetext{young}, the transitive adjective (i.e. a combination of adjective and preposition) \qacetext{mad-about}, the intransitive verb \qacetext{wait}, the transitive verb \qacetext{ask}, the adverb \qacetext{early}, the preposition \qacetext{for}, the variable \qacetext{X}, and the number \qacetext{2}.

We start with evaluating the ACE Codeco grammar. There are two properties we can test: (1) Is the language described by ACE Codeco unambiguous, and (2) is it indeed a subset of ACE?

\paragraph{Ambiguity Check of ACE Codeco.}

Languages like ACE are designed to be unambiguous on the syntactic level. This means that every valid sentence must have exactly one syntax tree. Exhaustive language generation operates on the syntax trees of sentences, which means that ambiguous sentences are generated several times with different syntax trees attached. Looking for duplicates in the list of generated sentences reveals whether there is ambiguity or not (for the given grammar, lexicon, and sentence length).

Up to the length of ten tokens, the evaluation subset of ACE Codeco generates 2\,250\,869 sentences. If sorted alphabetically, these are the first and last sentences:
\begin{quote}
\acetext{a  woman  asks  a  woman  .}\\
\acetext{a  woman  asks  a  woman  and  asks  a  woman  .}\\
\acetext{a  woman  asks  a  woman  and  asks  everybody  .}\\
\acetext{a  woman  asks  a  woman  and  asks  everybody  early  .}\\
...\\
\acetext{X  waits  for  X  for  X  for  who  ?}\\
\acetext{X  waits  for  X  for  X  for  who  early  ?}\\
\acetext{X  waits  for  X  for  X  for  X  .}\\
\acetext{X  waits  for  X  for  X  for  X  early  .}
\end{quote}
Due to the restricted lexicon, many of these sentences are repetitive. One should bear in mind that exhaustive language generation is not about generating natural sentences, but just syntactically well-formed ones. As it turns out, all generated sentences are distinct, which --- by design --- means that each has only one syntax tree. Thus, at least a large subset of ACE Codeco (its evaluation subset) is unambiguous for at least relatively short sentences (up to ten tokens).

\paragraph{Subset Check of ACE Codeco and Full ACE.}

The ACE Codeco grammar is designed as a proper subset of ACE. We can now check whether this is the case, again for the evaluation subset and up to a certain sentence length.

Every sentence up to the length of ten tokens --- as shown above --- was submitted to the ACE parser (APE) and parsing succeeded in all cases. Since APE is the reference implementation of ACE, this means that these sentences are syntactically correct ACE sentences.

\bigskip

\noindent Next, we can evaluate the parser implementation. There exists a second implementation of Codeco based on a simple transformation into Prolog DCGs \cite{kuhn2010doctoralthesis}, which can be used for comparison. This Prolog DCG implementation supports all Codeco elements, but does not implement lookahead features. We can therefore test the following two properties: (1) Are the two implementations equivalent, i.e. do they return the same results for given grammars, and (2) how fast are they compared to each other and compared to the ACE parser?

\paragraph{Equivalence Check of the Implementations.}

Both Codeco implementations (chart parser and Prolog DCG) support exhaustive language generation. This allows us to check whether the two implementations accept the same set of sentences, as they should. We take the ACE Codeco grammar to run this comparison.

Generating all sentences of ACE Codeco up to eight tokens results in identical sets of sentences for the two implementations. This is an indication that they contain no major bugs and interpret Codeco grammars in the same way.

\paragraph{Performance Tests of the Implementations.}

Finally, we can look at the performance of the two implementations. Both can be used for parsing and for generation, and thus the runtimes in these two disciplines can be compared. The first task was to generate all sentences of the evaluation subset up to the length of seven tokens. The second task was to parse the sentences that result from the generation task. This parsing task was performed in two ways for both implementations: once using the evaluation subset and once using the full ACE Codeco grammar. The restricted lexicon of the evaluation subset was used in both cases. These tests were performed on a MacBook Pro laptop computer having a 2.4 GHz Intel Core 2 Duo processor and 2 GB of main memory. Table~\ref{tab:performance} shows the results.
\begin{table}[t]
\begin{center}
\caption{The results of a performance test of the two implementations of Codeco, with the existing ACE parser (APE) for comparison (all numbers are rounded to three significant digits).}
\label{tab:performance}
\begin{tabular}{l@{~~}l@{~~}l@{~}|@{~}r@{.}l@{~~}r@{.}l}
& & & \multicolumn{4}{c}{time in seconds} \\
task & grammar & implementation & \multicolumn{2}{c}{\emph{overall}} & \multicolumn{2}{c}{\emph{average}} \\
\hline
generation & ACE Codeco eval. subset & Prolog DCG & 40&8 & 0&00286 \\
generation & ACE Codeco eval. subset & Java chart parser & 1040& & 0&0730 \\
parsing & ACE Codeco eval. subset & Prolog DCG & 5&13 & 0&000360 \\
parsing & ACE Codeco eval. subset & Java chart parser & 392& & 0&0276 \\
parsing & full ACE Codeco & Prolog DCG & 20&7 & 0&00146 \\
parsing & full ACE Codeco & Java chart parser & 1900& & 0&134 \\
parsing & full ACE & APE & 230& & 0&0161 \\
\hline
\end{tabular}
\end{center}
\end{table}

The generation of the resulting 14\,240 sentences only requires about 41 seconds in the case of the Prolog DCG implementation. This means that less than 3 milliseconds are needed on average for generating one sentence. The Java chart parser implementation needs about 17 minutes for this complete generation task, which corresponds to 73 milliseconds per sentence. Thus, generation is about 25 times faster when using the Prolog DCG version compared to the Java implementation. These results show that the Prolog DCG implementation is well suited for exhaustive language generation. The chart parser implementation is much slower but the time values are still within a time range that is more than reasonable.

The Prolog DCG approach is very fast for parsing the same set of sentences using the evaluation subset of the grammar. Here, parsing just means detecting that the given statements are well-formed according to the grammar. Altogether only slightly more than 5 seconds are needed to parse the complete test set, i.e. less than 0.4 milliseconds per sentence. When using the full ACE Codeco grammar for parsing the same set of sentences, altogether 21 seconds are needed, i.e. about 1.5 milliseconds per sentence. The chart parser implementation is again much slower and requires almost 30 milliseconds per sentence when using the grammar of the evaluation subset, which leads to an overall time of more than 6 minutes. For the full grammar, 134 milliseconds are required per sentence leading to an overall time of about 32 minutes. Thus, the chart parser implementation is 76 to 92 times slower than the Prolog DCG for the parsing task. Because all time values are clearly below 1 second per sentence, both parser implementations can be considered fast enough for practical applications.

The fact that the chart parser implementation in Java requires considerably more time than the Prolog DCG is not surprising. DCG grammar rules in Prolog are directly translated into Prolog clauses and generate only very little overhead. Java, in contrast, has no special support for grammar rules: they have to be implemented on a higher level. The same holds for variable unifications, which come for free with Prolog but have to be implemented on a higher level in Java.

As a comparison, the existing parser APE --- the reference implementation of ACE --- needs about 4 minutes for the complete parsing task. Thus, it is faster than the chart parser but slower than the Prolog DCG. However, it has to be considered that APE does more than just accepting well-formed sentences: It also generates logical representations and syntax trees.

\bigskip

\noindent In summary, the ACE Codeco grammar can be considered unambiguous and fully ACE-com\-pli\-ant. The chart parser implementation is stable and reasonably fast (even though much slower than a Prolog DCG). From this we can conclude that the Codeco notation is suitable for describing controlled languages like ACE. In addition, it shows that Codeco meets the implementability requirement introduced in Section \ref{sec:grammarrequ}.

\section{Conclusions}
\label{sec:conclusions}

With Codeco, controlled natural languages of the unambiguous type can be defined in a simple and convenient way. With the help of different kinds of forward and backward references, complex nonlocal structures like anaphoric references can be defined in a fully declarative way. Following a chart parsing approach, the possible continuations of unfinished sentences can be reliably retrieved, which facilitates the implementation of predictive editors. Implementing efficient parsers for Codeco is not difficult and does not depend on a particular programming paradigm. In addition, the presented approach enables automatic grammar testing, e.g. by exhaustive language generation, which can be considered very important for the development of reliable practical applications.

Codeco can be conceived as a proposal for a general CNL grammar notation. It is possible that extensions become necessary to cover other controlled natural languages, but Codeco has been shown to work very well for a large subset of ACE, which is one of the most advanced CNLs to date.


\bibliographystyle{spmpsci}
\bibliography{jlli_codeco}

\begin{thebibliography}{10}
\providecommand{\url}[1]{{#1}}
\providecommand{\urlprefix}{URL }
\expandafter\ifx\csname urlstyle\endcsname\relax
  \providecommand{\doi}[1]{DOI~\discretionary{}{}{}#1}\else
  \providecommand{\doi}{DOI~\discretionary{}{}{}\begingroup
  \urlstyle{rm}\Url}\fi

\bibitem{adriaens1992coling}
Adriaens, G., Schreors, D.: From {COGRAM} to {ALCOGRAM}: Toward a controlled
  english grammar checker.
\newblock In: Proceedings of the 14th Conference on Computational Linguistics,
  vol.~2, pp. 595--601. Association for Computational Linguistics, Morristown,
  NJ, USA (1992)

\bibitem{angelov2009cnlmain}
Angelov, K., Ranta, A.: Implementing controlled languages in {GF}.
\newblock In: Proceedings of the Workshop on Controlled Natural Language ({CNL}
  2009), \emph{Lecture Notes in Computer Science}, vol. 5972, pp. 82--101.
  Springer (2010)

\bibitem{bernstein2006iswc}
Bernstein, A., Kaufmann, E.: {GINO} --- a guided input natural language
  ontology editor.
\newblock In: The {S}emantic {W}eb --- {ISWC} 2006, Proceedings of the 5th
  International {S}emantic {W}eb Conference, \emph{Lecture Notes in Computer
  Science}, vol. 4273, pp. 144--157. Springer (2006)

\bibitem{chomsky1980inquiry}
Chomsky, N.: On binding.
\newblock Linguistic Inquiry \textbf{11}(1), 1--46 (1980)

\bibitem{clark2007kcap}
Clark, P., Chaw, S.Y., Barker, K., Chaudhri, V., Harrison, P., Fan, J., John,
  B., Porter, B., Spaulding, A., Thompson, J., Yeh, P.: Capturing and answering
  questions posed to a knowledge-based system.
\newblock In: {K-CAP} '07: Proceedings of the 4th International Conference on
  Knowledge Capture, pp. 63--70. ACM (2007)

\bibitem{clark2005flairs}
Clark, P., Harrison, P., Jenkins, T., Thompson, J., Wojcik, R.H.: Acquiring and
  using world knowledge using a restricted subset of {E}nglish.
\newblock In: Proceedings of the Eighteenth International Florida Artificial
  Intelligence Research Society Conference ({FLAIRS} 2005), pp. 506--511. AAAI
  Press (2005)

\bibitem{cole1997hlt}
Cole, R., Mariani, J., Uszkoreit, H., Varile, G.B., Zaenen, A., Zampolli, A.,
  Zue, V. (eds.): Survey of the State of the Art in Human Language Technology.
\newblock Cambridge University Press (1997)

\bibitem{covington1994nlpprolog}
Covington, M.A.: Natural Language Processing for {P}rolog Programmers.
\newblock Prentice Hall, Englewood Cliffs, NJ, USA (1994)

\bibitem{dahl1997iclp}
Dahl, V., Tarau, P., Li, R.: Assumption grammars for processing natural
  language.
\newblock In: L.~Naish (ed.) Proceedings of the Fourteenth International
  Conference on Logic Programming, pp. 256--270. MIT Press (1997)

\bibitem{dimitrova2008iswc}
Dimitrova, V., Denaux, R., Hart, G., Dolbear, C., Holt, I., Cohn, A.G.:
  Involving domain experts in authoring owl ontologies.
\newblock The Semantic Web --- Proceedings of the 7th International Semantic
  Web Conference (ISWC 2008) pp. 1--16 (2008)

\bibitem{earley1970acm}
Earley, J.: An efficient context-free parsing algorithm.
\newblock Communications of the ACM \textbf{13}(2), 94--102 (1970)

\bibitem{franconi2011dl}
Franconi, E., Guagliardo, P., Tessaris, S., Trevisan, M.: Quelo: an
  ontology-driven query interface.
\newblock In: Proceedings of the 24th International Workshop on Description
  Logics (DL 2011) (2011)

\bibitem{fuchs2008reasoningweb}
Fuchs, N.E., Kaljurand, K., Kuhn, T.: {A}ttempto {C}ontrolled {E}nglish for
  knowledge representation.
\newblock In: Reasoning {W}eb --- 4th International Summer School 2008,
  \emph{Lecture Notes in Computer Science}, vol. 5224, pp. 104--124. Springer
  (2008)

\bibitem{fuchs1999lopstr}
Fuchs, N.E., Schwertel, U., Schwitter, R.: Attempto controlled english - not
  just another logic specification language.
\newblock In: Proceedings of the 8th International Workshop on Logic
  Programming Synthesis and Transformation (LOPSTR '98). Springer-Verlag (1990)

\bibitem{funk2007iswc}
Funk, A., Tablan, V., Bontcheva, K., Cunningham, H., Davis, B., Handschuh, S.:
  {CLOnE}: Controlled language for ontology editing.
\newblock In: Proceedings of the 6th International {S}emantic {W}eb Conference
  and the 2nd Asian {S}emantic {W}eb Conference ({ISWC} 2007 + {ASWC} 2007),
  \emph{Lecture Notes in Computer Science}, vol. 4825, pp. 142--155. Springer
  (2007)

\bibitem{gazdar1985gpsg}
Gazdar, G.: Generalized Phrase Structure Grammar.
\newblock Harvard University Press (1985)

\bibitem{gazdar1989prolognlp}
Gazdar, G., Mellish, C.: Natural Language Processing in {PROLOG}.
\newblock Addison-Wesley (1989)

\bibitem{grune2008parsing}
Grune, D., Jacobs, C.J.: Parsing Techniques --- A Practical Guide, second edn.
\newblock Monographs in Computer Science. Springer Science+Business Media, New
  York, NY, USA (2008)

\bibitem{hobbs1978lingua}
Hobbs, J.R.: Resolving pronoun references.
\newblock Lingua \textbf{44}(4), 311--338 (1978)

\bibitem{johnson1986coling}
Johnson, M., Klein, E.: Discourse, anaphora and parsing.
\newblock In: Proceedings of the 11th coference on Computational linguistics,
  COLING '86, pp. 669--675. Association for Computational Linguistics,
  Stroudsburg, PA, USA (1986)

\bibitem{johnson1975yacc}
Johnson, S.C.: {Y}acc: Yet another compiler-compiler.
\newblock Computer Science Technical Report~32, Bell Laboratories, Murray Hill,
  NJ, USA (1975)

\bibitem{joshi1975computsystsci}
Joshi, A.K., Levy, L.S., Takahashi, M.: Tree adjunct grammars.
\newblock Journal of Computer and System Sciences \textbf{10}(1), 136--163
  (1975)

\bibitem{kaplan1982mental}
Kaplan, R.M., Bresnan, J.: Lexical-functional grammar: A formal system for
  grammatical representation.
\newblock In: J.~Bresnan (ed.) The Mental Representation of Grammatical
  Relations, pp. 173--281. MIT Press (1982)

\bibitem{knuth1964acm}
Knuth, D.E.: Backus normal form vs. backus naur form.
\newblock Communications of the ACM \textbf{7}(12), 735--736 (1964)

\bibitem{kuhn2007rr}
Kuhn, T.: {AceRules}: Executing rules in controlled natural language.
\newblock In: {W}eb Reasoning and Rule Systems --- First International
  Conference ({RR} 2007), \emph{Lecture Notes in Computer Science}, vol. 4524,
  pp. 299--308. Springer (2007)

\bibitem{kuhn2008swui}
Kuhn, T.: {AceWiki}: A natural and expressive semantic wiki.
\newblock In: Proceedings of the Fifth International Workshop on Semantic Web
  User Interaction (SWUI 2008) --- Exploring HCI Challenges, \emph{CEUR
  Workshop Proceedings}, vol. 543. CEUR-WS (2009).
\newblock \urlprefix\url{http://ceur-ws.org/Vol-543/kuhn_swui2008.pdf}

\bibitem{kuhn2009semwiki}
Kuhn, T.: How controlled {E}nglish can improve semantic wikis.
\newblock In: Proceedings of the Forth Semantic Wiki Workshop ({SemWiki} 2009),
  \emph{CEUR Workshop Proceedings}, vol. 464. CEUR-WS (2009)

\bibitem{kuhn2010doctoralthesis}
Kuhn, T.: {Controlled English for Knowledge Representation}.
\newblock Ph.D. thesis, {Faculty of Economics, Business Administration and
  Information Technology of the University of Zurich} (2010)

\bibitem{kuhn2010cnlmain}
Kuhn, T.: {C}odeco: A practical notation for controlled {E}nglish grammars in
  predictive editors.
\newblock In: Proceedings of the Second Workshop on Controlled Natural Language
  ({CNL} 2010), \emph{Lecture Notes in Computer Science}, vol. 7175, pp.
  95--114. Springer (2012)

\bibitem{kuhn2012swj}
Kuhn, T.: The understandability of {OWL} statements in controlled {E}nglish.
\newblock Semantic Web journal  (to appear)

\bibitem{kuhnhoefler2013coral}
Kuhn, T., Hoefler, S.: {C}oral: Corpus access in controlled language.
\newblock Corpora \textbf{7}(2) (2012, to appear)

\bibitem{kuhn2008alta}
Kuhn, T., Schwitter, R.: Writing support for controlled natural languages.
\newblock In: Proceedings of the Australasian Language Technology Association
  Workshop 2008, pp. 46--54 (2008)

\bibitem{lappin1994coli}
Lappin, S., Leass, H.J.: An algorithm for pronominal anaphora resolution.
\newblock Computational Linguistics \textbf{20}(4), 535--561 (1994)

\bibitem{martin2002iccs}
Martin, P.: Knowledge representation in {CGLF}, {CGIF}, {KIF}, {Frame-CG} and
  {F}ormalized-{E}nglish.
\newblock In: Conceptual Structures: Integration and Interfaces --- Proceedings
  of the 10th International Conference on Conceptual Structures ({ICCS} 2002),
  \emph{Lecture Notes in Artificial Intelligence}, vol. 2393, pp. 77--91.
  Springer (2002)

\bibitem{mueckstein1985csc}
Mueckstein, E.M.: {Controlled natural language interfaces: the best of three
  worlds}.
\newblock In: {CSC} '85: Proceedings of the 1985 {ACM} thirteenth annual
  conference on Computer Science, pp. 176--178. ACM (1985)

\bibitem{naur1963acm}
Naur, P., Backus, J.W., Bauer, F.L., Green, J., Katz, C., McCarthy, J., Perils,
  A.J., Rutishauser, H., Samelson, K., Vauquois, B., Wegstein, J.H., van
  Wijngaarden, A., Woodger, M.: Revised report on the algorithmic language
  {ALGOL} 60.
\newblock Communications of the ACM \textbf{6}(1), 1--17 (1963)

\bibitem{ogden1932basic}
Ogden, C.K.: The {A} {B} {C} of {B}asic {E}nglish (in {B}asic).
\newblock No.~43 in Psyche Miniatures General Series. K. Paul, Trench, Trubner,
  London (1932)

\bibitem{pereira1986nlp}
Pereira, F., Warren, D.H.D.: Definite clause grammars for language analysis.
\newblock In: Readings in Natural Language Processing, pp. 101--124. Morgan
  Kaufmann Publishers (1986)

\bibitem{pollard1994hpsg}
Pollard, C., Sag, I.: Head-Driven Phrase Structure Grammar.
\newblock Studies in Contemporary Linguistics. Chicago University Press (1994)

\bibitem{pool2006claw}
Pool, J.: Can controlled languages scale to the {W}eb?
\newblock In: Proceedings of the 5th International Workshop on Controlled
  Language Applications ({CLAW} 2006) (2006)

\bibitem{power2009cnl}
Power, R., Stevens, R., Scott, D., Rector, A.: Editing {OWL} through generated
  {CNL}.
\newblock In: Pre-Proceedings of the Workshop on Controlled Natural Language
  ({CNL} 2009), \emph{CEUR Workshop Proceedings}, vol. 448. CEUR-WS (2009)

\bibitem{schwitter2008owleddc}
Schwitter, R., Kaljurand, K., Cregan, A., Dolbear, C., Hart, G.: A comparison
  of three controlled natural languages for {OWL} 1.1.
\newblock In: Proceedings of the Fourth {OWLED} Workshop on {OWL}: Experiences
  and Directions, \emph{CEUR Workshop Proceedings}, vol. 496. CEUR-WS (2008)

\bibitem{schwitter2003eamtclaw}
Schwitter, R., Ljungberg, A., Hood, D.: {ECOLE} --- a look-ahead editor for a
  controlled language.
\newblock In: Controlled Translation --- Proceedings of the Joint Conference
  combining the 8th International Workshop of the European Association for
  Machine Translation and the 4th Controlled Language Application Workshop
  ({EAMT-CLAW03}), pp. 141--150. Dublin City University, Ireland (2003)

\bibitem{shiffman2009cnlmain}
Shiffman, R.N., Michel, G., Krauthammer, M., Fuchs, N.E., Kaljurand, K., Kuhn,
  T.: Writing clinical practice guidelines in controlled natural language.
\newblock In: Proceedings of the Workshop on Controlled Natural Language ({CNL}
  2009), \emph{Lecture Notes in Computer Science}, vol. 5972, pp. 265--280.
  Springer (2010)

\bibitem{spreeuwenberg2009cnlmain}
Spreeuwenberg, S., Anderson~Healy, K.: {SBVR}'s approach to controlled natural
  language.
\newblock In: Proceedings of the Workshop on Controlled Natural Language ({CNL}
  2009), \emph{Lecture Notes in Computer Science}, vol. 5972, pp. 155--169.
  Springer (2010)

\bibitem{steedman2011ccg}
Steedman, M., Baldridge, J.: Combinatory categorial grammar.
\newblock In: Non-Transformational Syntax, pp. 181--224. Wiley-Blackwell (2011)

\bibitem{pulman1999iwcs}
Sukkarieh, J.Z., Pulman, S.G.: {C}omputer {P}rocessable {E}nglish and
  {M}c{L}ogic.
\newblock In: Proceedings of the Third International Workshop on Computational
  Semantics, pp. 367--380 (1999)

\bibitem{tennant1983acl}
Tennant, H.R., Ross, K.M., Saenz, R.M., Thompson, C.W., Miller, J.R.:
  Menu-based natural language understanding.
\newblock In: Proceedings of the 21st annual meeting on Association for
  Computational Linguistics, pp. 151--158. Association for Computational
  Linguistics (1983)

\bibitem{verbeke1973traindev}
Verbeke, C.A.: {C}aterpillar fundamental {E}nglish.
\newblock Training \& Development Journal \textbf{27}(2), 36--40 (1973)

\bibitem{wuersch2010icse}
W\"{u}rsch, M., Ghezzi, G., Reif, G., Gall, H.C.: Supporting developers with
  natural language queries.
\newblock In: ICSE '10: Proceedings of the 32nd ACM/IEEE International
  Conference on Software Engineering, pp. 165--174. ACM, New York, NY, USA
  (2010)

\bibitem{wyner2009cnlmain}
Wyner, A., Angelov, K., Barzdins, G., Damljanovic, D., Davis, B., Fuchs, N.,
  Hoefler, S., Jones, K., Kaljurand, K., Kuhn, T., Luts, M., Pool, J., Rosner,
  M., Schwitter, R., Sowa, J.: On controlled natural languages: Properties and
  prospects.
\newblock In: Proceedings of the Workshop on Controlled Natural Language ({CNL}
  2009), \emph{Lecture Notes in Computer Science}, vol. 5972, pp. 281--289.
  Springer (2010)

\end{thebibliography}

\end{document}